\documentclass{article}

\usepackage[preprint]{neurips_2023}

\usepackage[utf8]{inputenc} 
\usepackage[T1]{fontenc}    
\usepackage{hyperref}       
\usepackage{url}            
\usepackage{booktabs}       
\usepackage{amsfonts}       
\usepackage{nicefrac}       
\usepackage{microtype}      
\usepackage{xcolor}         

\usepackage{multirow}
\usepackage[normalem]{ulem}
\usepackage[linesnumbered,ruled]{algorithm2e}
\usepackage{amsmath}
\usepackage{cleveref}
\usepackage{booktabs}
\usepackage{varwidth}
\usepackage{paralist}
\usepackage{enumerate}
\usepackage{amssymb}
\usepackage{pifont}
\usepackage{bm}
\usepackage{amsmath}
\usepackage{multirow}
\usepackage{caption}
\usepackage{graphicx}
\usepackage{subfigure}

\crefname{equation}{Eq.}{Eq.}
\crefname{section}{Section}{Sections}
\crefname{subsection}{Section}{Sections}
\crefname{subsubsection}{Section}{Sections}
\crefname{figure}{Figure}{Figures}
\crefname{table}{Table}{Tables}
\crefname{subfigure}{Figure}{Figures}
\crefname{algocf}{Algorithm}{Algorithms}

\newcommand{\xmark}{\ding{55}}%

\newcommand{\Trans}[1]{{#1}^{\top}}

\title{Robust and Interpretable Medical Image Classifiers via Concept Bottleneck Models}

\author{
  An Yan$^{\diamondsuit}$,
  Yu Wang$^{\diamondsuit}$,
  Petros Karypis$^{\diamondsuit}$,
  Zexue He$^{\diamondsuit}$,
  Chengyu Dong$^{\diamondsuit}$,
  Zihan Wang$^{\diamondsuit}$,
  Yiwu Zhong$^{\clubsuit}$, 
  \\
  \textbf{
  Jingbo Shang$^{\diamondsuit}$, 
  Amilcare Gentili$^{\diamondsuit}$,
  Chun-Nan Hsu$^{\diamondsuit\spadesuit}$,
  Julian McAuley$^{\diamondsuit\spadesuit}$}
  \\
  $^{\diamondsuit}$UC San Diego, 
  $^{\clubsuit}$University of Wisconsin-Madison,
  $^{\spadesuit}$ VA San Diego Healthcare System
  \\
  {\tt \small \{ayan,yuw164,zehe,cdong,ziw224,agentili,chunnan,jshang,jmcauley\}@ucsd.edu}
  {\tt \small yzhong52@wisc.edu}
}

\begin{document}

\maketitle

\begin{abstract}
  Medical image classification is a critical problem for healthcare, with the potential to alleviate the workload of doctors and facilitate diagnoses of patients. However, two challenges arise when deploying deep learning models to real-world healthcare applications.
  First, neural models tend to learn spurious correlations instead of desired features, which could fall short when generalizing to new domains (e.g., patients with different ages).
  Second, these black-box models lack interpretability. When making diagnostic predictions, it is important to understand why a model makes a decision for trustworthy and safety considerations.
  In this paper, to address these two limitations, we propose a new paradigm to build robust and interpretable medical image classifiers with natural language concepts. Specifically, we first query clinical concepts from GPT-4, then transform latent image features into explicit concepts with a vision-language model.
  We systematically evaluate our method on eight medical image classification datasets to verify its effectiveness. On challenging datasets with strong confounding factors, our method can mitigate spurious correlations thus substantially outperform standard visual encoders and other baselines. Finally, we show how classification with a small number of concepts brings a level of interpretability for understanding model decisions through case studies in real medical data.

\end{abstract}


\section{Introduction}
Medical image classification is a critical yet challenging problem in machine learning for healthcare. The development of deep learning models has demonstrated great success~\cite{litjens2017survey,azizi2021big,shamshad2023transformers,lu2022semi}, by achieving superior performance in benchmarks and competitions. However, there are two unsolved problems that still prevent us from deploying these models in clinical usages.

The first problem is the presence of confounding factors~\citep{zech2018confounding,santa2021public,wilds} in medical data that hurts generalization. Neural networks are prone to learn spurious correlations for classification tasks. 
In the non-medical domain, \cite{sagawa2020investigation} found that models trained on the Waterbirds dataset correlate waterbirds with backgrounds containing water, and models trained on the CelebA dataset~\citep{liu2018large} correlate males with dark hair.
This could be more of an issue for medical image classification, as confounding factors broadly exist and labeled data are often limited~\citep{de2016machine}. 
Take the classification of patient X-rays between Covid-19 and normal for instance, certain factors such as the hospitals where the X-rays are performed and the age of the patient strongly correlate with the target disease classification. 
To quantify this issue, we curated datasets of known confounding factors such as hospitals, age and gender, and found that standard visual classifiers and previous popular methods designed to mitigate spurious correlations often perform poorly and struggle to generalize in these datasets. As a concrete example, instead of learning to predict Covid or normal, the classifier might instead learn to predict if the X-ray is from a young or old patient. 

The second problem is the lack of interpretability. Deep neural networks are inherently ``black-box'' models due to their complex non-linear structures. This will raise safety and trust issues, as it is hard for human to understand model behaviors and trust model decisions at ease. It is especially important in clinical settings~\citep{rudin2022black}, as the adoption of deep learning models relies on building trust with healthcare professionals and patients. Clinicians often need to understand the underlying reasoning of the models to carefully make their decisions. Interpretable medical image classification models~\citep{clough2019global,shen2021interpretable,thomas2021interpretable,van2022explainable} allow for better error analysis, bias detection, ensuring patients safety, and trust building.

\begin{figure}[t]
  \centering 
  \includegraphics[width=1\linewidth]{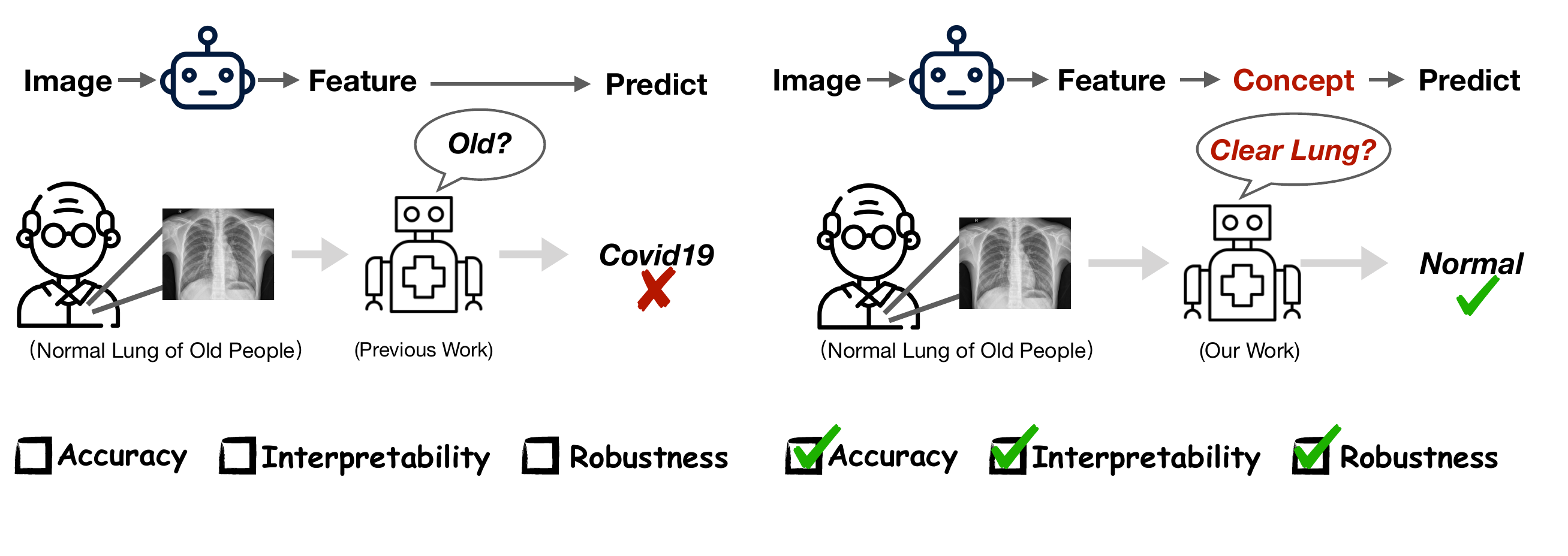}  
  \vspace{-30pt}
  \caption{High level illustration of our new paradigm. It utilizes concepts for medical image classification to achieve interpretability and robustness while maintaining accuracy. \textbf{Left}: Classification with a visual encoder; \textbf{Right}: Classification with concepts. A Chest X-ray from a healthy old individual may be classified as Covid-19 because of the age, while our method can mitigate spurious correlation by classifying with clinical concepts.
  }
  \label{fig:teaser}
  \vspace{-5pt}
\end{figure}

But how do we build a system that is both robust and interpretable? Inspired by recent work that uses concepts~\citep{koh2020concept,yuksekgonul2022post} or descriptions~\citep{menon2022visual} to amplify image classification and gain interpretation, in this paper, we address the two clinical challenges in a unified framework through natural language concepts ( illustrated in~\cref{fig:teaser}.).
Specifically, we elicit medical knowledge from large language models (e.g., GPT-4) in a zero-shot manner to build a set of concepts, i.e., concise descriptors regarding each disease or pathology, and project visual features into the concept space using a vision-language model to connect two modalities, and finally classify medical images with the concept vector. 
By doing so, we explicitly tell the model to learn desired features rather than possible spurious correlations, hence improving robustness while gaining interpretation.

We conduct experiments on eight datasets with case studies and human evaluation, and find several advantages with this new paradigm: 
\begin{enumerate}
    \item  It is easy to build in an automatic way, with minimal human effort and medical expertise. 
    \item On challenging datasets with strong confounding factors, classification using concepts can alleviate spurious correlations and substantially improve classification performance: an average of 19\% accuracy improvement over using raw image features.
    \item Even on popular benchmarks without explicit confounding factors, where the train and test set distributions are assumed to be the same, this new paradigm still attain competitive and sometimes even better performance than black-box visual encoders.
    \item Moreover, we gain a level of interpretability through this framework, by associating images with a small number of relevant concepts contributing to classification.
\end{enumerate}
\vspace{-5pt}


\section{Methodology}
\begin{figure}[t]
  \centering 
  \includegraphics[width=0.95\linewidth]{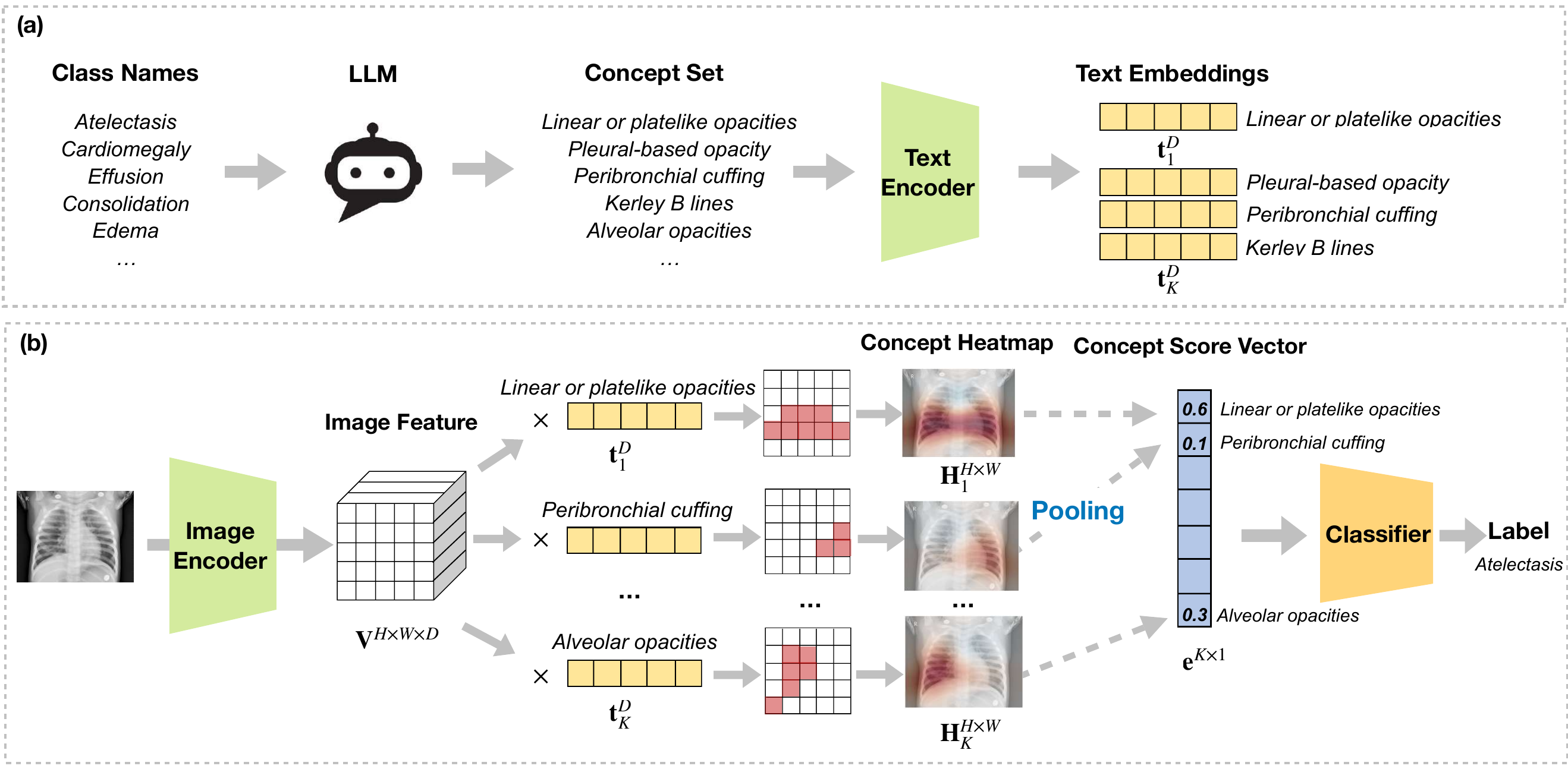}  
  \caption{The overall framework of classification with concepts. (a) Eliciting medical knowledge from GPT-4. (b) Projecting visual features into the concept space for classification.
  }
  \label{fig:model}
  \vspace{-10pt}
\end{figure}
In this section, we present the key components for building robust and interpretable medical image classifiers via natural language concepts.
First, in~\cref{sec:method_llm}, we interact with a large language model, GPT-4, to generate useful medical concepts.
Second, in~\cref{sec:method_clip}, we show how to effectively project visual features into concept embeddings along with in-depth analysis.
Finally, in~\cref{sec:method_fc}, we learn a classification layer and explain the interpretability of our model by design. An illustration of the framework is shown in~\cref{fig:model}.

\subsection{Eliciting Medical Knowledge from Large Language Models}
\label{sec:method_llm}
Given $M_c$ classes of target diseases or pathology, the first step of our paradigm is to acquire a set of useful concepts related to the classes. 
A typical workflow in healthcare and medical domain is to seek help from experts. 
We show that the recent instruction-following large language models, such as GPT-4, present a new alternative to solve language-related problems in healthcare~\citep{nori2023capabilities}. 
Specifically, given a disease (e.g., Atelectasis) we start by prompting the model using the following question to query for knowledge:\\
\textit{Q: What are the useful visual attributes to recognize \{Atelectasis\} in a chest X-ray?}

As large language models also present the ability to reason~\citep{wei2022chain,kojima2022large}, we can further prompt the model with a group of classes and ask it to reason about discriminative visual features to classify a number of classes:\\
\textit{Q: What are the useful visual attributes to distinguish \{Atelectasis, Effusion, ...\} in a chest X-ray?}

Recent work that query LLMs~\citep{oikarinen2023label} to acquire concepts falls into the query-then-filter schema, where they rely on post-processing and filtering with manually-designed rules.
We simplify these procedures by leveraging GPT-4 as a versatile tool for advanced clinical natural language processing, e.g., revising radiology descriptors to be concise, or removing the redundancy in text and finding a set of distinctive $N$ concepts given any arbitrary number $N$. An example query could be:\\
\textit{Q: Here is a list of concepts. Can you select the most distinctive 5 concepts from them?}

\begin{figure}[t]
  \centering 
  \includegraphics[width=1\linewidth]{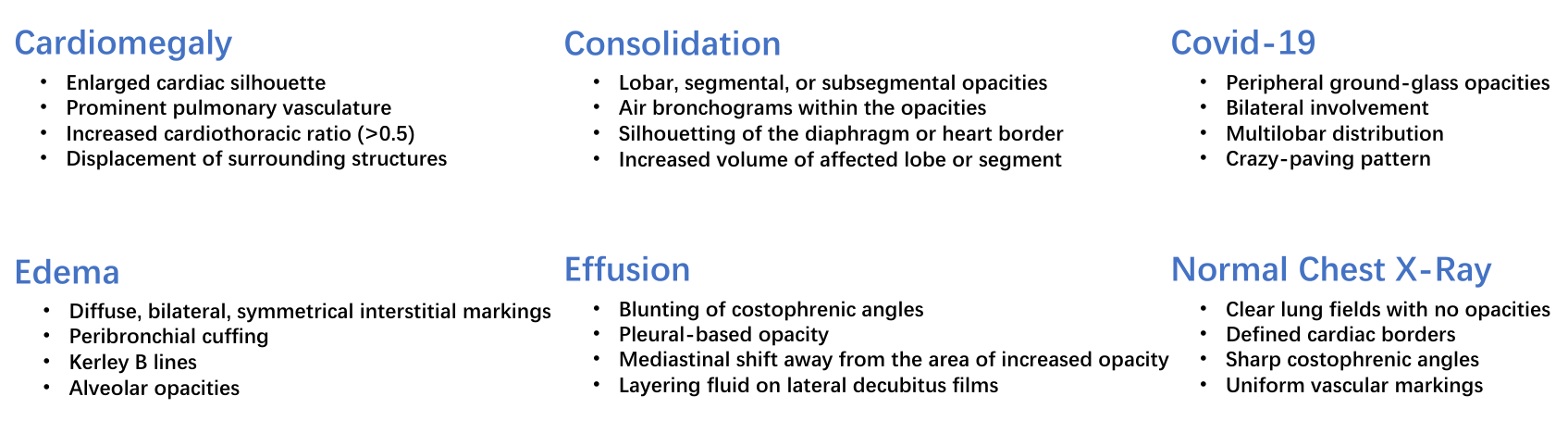}  
  \caption{Examples of concepts produced by GPT-4.
  }
  \label{fig:example-concepts}
  \vspace{-10pt}
\end{figure}

We present more details of interactions with GPT-4 and robustness checks for prompts in~\cref{app:prompting}.
Examples of medical concepts queried from GPT-4 are presented in~\cref{fig:example-concepts}.

\subsection{Mapping Visual Features into a Concept Space}
\label{sec:method_clip}
After obtaining a set of $N$ useful concepts $\mathcal{C} = \{c_1, c_2, \ldots, c_N\}$ from GPT-4, the next challenge is how to effectively connect those concepts with medical images to build interpretable visual classifiers. 
Previous work~\citep{koh2020concept} requires expert annotations to label the correlation between an image and a concept to enable the transformation, which is time-consuming and costly. 
The recent advance of vision-language models~(VLM), that are contrastively pre-trained with large-scale image and text pairs, present a new opportunity to automatically align images with texts.

In this paper, we leverage a VLM specialized in medical domain, BioViL~\citep{boecking2022making}, to connect medical concepts with images.  
Given an image $I$ and a concept set $\mathcal{C} = \{c_1, c_2, \ldots, c_N\}$, we extract its feature map $\textbf{V} \in \mathbb{R}^{H\times W\times D}$ and the text embeddings $\textbf{t}_i \in R^D$ as follows:
\begin{align}
    \textbf{V} &= \Theta_V(I), \\ 
    \textbf{t}_i &= \Theta_T(a_i), i=1, \ldots, N  
\end{align}
where $\Theta_V$ and $\Theta_T$ are the two stream visual and text encoders in a VLM, $\textbf{t}_i$ is the embedding of the $i$-th concept in the concept pool, $H$ and $W$ are the height and width of the feature map. 

Given $\textbf{V}$ and $\textbf{t}_i$, one can obtain a heatmap $\textbf{H}_i$, i.e., a similarity matrix measuring the similarity between the concept and the image, by computing their cosine distance:
\begin{equation}
    \textbf{H}_i^{j,k} = \frac{\Trans{\textbf{t}}_i \textbf{V}_{j,k}}{||\textbf{t}_i|| || \textbf{V}_{j,k}||}, j=1,\ldots,H, k=1,\ldots,W
\end{equation}
where $j$, $k$ are the j-th and k-th position in the row and column, and $\textbf{H}_i^{j,k}$ represent a local similarity score between $\textbf{V}$ and $\textbf{t}_i$.

Then the problem of connecting image and text is simplified as learning from the set of heatmaps $\{H_1, H_2, \ldots, H_N\}$.

A simple and intuitive approach is to apply average or max pooling given the heatmaps:
\begin{equation}
    s_i  = \frac{1}{H \cdot W}\sum_{j=1}^H \sum_{k=1}^W \textbf{H}_i^{j,k} 
\end{equation}

\begin{equation}
    s_i  = {H}_i^{j^*,k^*} = \max (\textbf{H}_i^{j,k} ).
\end{equation}

Though more advanced methods such as convolution, self-attention, or adding positional embeddings are possible, empirically we found average pooling works well for downstream classification (see ablations in Appendix~\cref{tab:ablation_conv}), especially for challenging datasets with significant distribution shift between training and testing. Moreover, since no additional training is involved in this step, we can treat those scores as a continuous measurement of whether a concept is related to the image, given a pre-trained vision-language model.

In the end, we obtain a concept vector $\textbf{e}$, representing the similarity between an image and a set of concepts:
\begin{align}
    \textbf{e} &= (s_1, \ldots, s_N)^T.
\end{align}

\subsection{Learning a Classification Layer}
\label{sec:method_fc}

Our final step is to learn a decision layer to predict $M_c$ classes. To preserve interpretability, we normalize concept vectors $e$ to a scale of 0 and 1, and use a fully connected layer 
$\textbf{W}^F \in \mathbb{R}^{ M_c \times N}$ without bias term for learning and inference.

Given a concept vector $e_i$ denoting the i-th image in a batch, we project the concept scores into logits and then use softmax to obtain the probabilities:
\begin{align}
    \textbf{z}_i &= \textbf{W}^F \cdot \textbf{e}_i, \\
    \textbf{p}_i &= \text{softmax}(Z) = \frac{e^{\textbf{z}_i}}{\sum_{c=1}^{M_c} e^{z_{i,c}}}.
\end{align}
    
We use the categorical cross-entropy loss for training a classification model:
\begin{equation}\label{eq:loss_ce}
    \mathcal{L}_{\mathit{ce}} = -\frac{1}{K} \sum_{i=1}^K \sum_{c=1}^{M_C} y_{i,c} \log(p_{i,c}),
\end{equation}
where $K$ is the number of images in a mini-batch, $y_{i,c}$ is the binary indicator of the $i$-th image belonging to class $c$, and $p_{i,c}$ is the predicted probability of the $i$-th image belonging to class $c$.

Since each logit $z_{i,c}$ (corresponding to $p_{i,c}$) is a linear combination of non-negative concept scores $\textbf{e}_i = (s_1^{(i)}, \ldots, s_N^{(i)})^T$, i.e., $z_{i,c} = \sum_{j=1}^N \textbf{W}^F_{j,c} s_j^{(i)}$. We can interpret the weights in $W_F$ as importance scores for classifying class $c$ with concepts. Higher absolute values in ${W}^F_{j,c}$ indicate that the concept is more important for classification. Negative weights can be interpreted as negations, i.e., the non-existence of this concept. Overall, this final linear layer offers a way for human to better understand and analyze model decisions, as shown in~\cref{sec:exp_interpretable}.

\section{Experiments}
In this section, we first introduce datasets and baselines in~\cref{sec:exp_setup}, then validate the robustness and interpretability of our method in~\cref{sec:exp_robust} and~\ref{sec:exp_interpretable}, along with ablation studies in~\cref{sec:exp_ablation}.

\subsection{Experimental Setup}
\label{sec:exp_setup}
\paragraph{Datasets} We conduct experiments on eight datasets with two different settings. 

First, we create four datasets with strong confounding factors as follows: \textbf{(1) NIH-gender}: We use the meta data from NIH-CXR~\citep{wang2017chestx} to build a subset with male and female chest X-rays for classifying Atelectasis and Effusion. The training set consists of male patients with Atelectasis and females with Effusion, and the test set consists of female patients with Atelectasis and males with Effusion. The confounding factor here is gender.
\textbf{(2) NIH-age}: We build a subset from NIH-CXR with young and old chest X-rays to classify normal and abnormal. The training set consists of normal X-rays from young patients (age $\leq$ 16), and abnormal X-rays from old patients (age $\geq$ 70); the test set has the opposite combination. The confounding factor for this dataset is age.
\textbf{(3) NIH-agemix}: Similar to NIH-age, the normal cases in the training set consist of 90\% young patients and 10\% old paitents, while the abnormal cases consist of 90\% old patients and 10\% young patients.
\textbf{(4) Covid-mix}: We create a dataset from various sources to classify Covid-19 and non-Covid Pneumonia. The Covid-19 samples are from a collection of open source releases~\citep{cohen2020covid}, including Italian Society of Radiology~\citep{harmon2020artificial}, etc. This results in 2,313 Covid X-rays in total. We build a balanced dataset by providing 2,313 penumonia X-rays from two medical systems for training~\citep{kermany2018identifying} and testing~\cite{wang2017chestx}.
There can be several confounding factors in this dataset, for example, hospital, age and gender. 

Second, we evaluate models on four standard benchmarks. 
These datasets do not have explicit confounding factors, since the training and test samples are assumed to be randomly drawn from the same distribution. These datasets include: 
\textbf{(1) NIH-CXR}~\citep{wang2017chestx}: A large public dataset of chest X-rays, with 108,948 front view Chest x-ray images of 32,717 patients collected from NIH Clinical Center.
\textbf{(2) Covid-QU}~\citep{chowdhury2020can}: A Covid dataset collected by Qatar University along with normal, lung opacity and pneumonia.
\textbf{(3) Pneumonia}~\citep{kermany2018identifying}: A public dataset for detecting pneumonia. Its chest X-rays were selected from pediatric patients of one to five years old in Guangzhou Women and Children’s Medical Center. 
\textbf{(4) Open-i}~\citep{demner2012design}: Chest X-rays collected by Indiana University Hospital through the network from open source literature and biomedical image collection, with 7,470 frontal and lateral chest films. 

\paragraph{Baselines} We compare our method with three types of baselines: domain generalization models designed to mitigate spurious correlations, visual encoders, and two recent methods that build Concept Bottleneck Models (CBMs) for image classification. 
\begin{itemize}
\item ERM~\citep{shi2021gradient}. A baseline method that trains an end-to-end visual encoder with Empirical Risk Minimization.
    \item Fish~\citep{shi2021gradient}. Using inter-domain gradient matching objective to maximize the similarity between the gradients from different domains for generalization.
    \item LISA~\citep{lisa}. A recent method utilizing intra-domain and intra-label mixup to mitigate the effects of the confounders and achieve better out-of-domain performance. 
    \item BioViL Image Features. Extracting image features of BioViL~\citep{boecking2022making} for linear probing. We also consider adding dropouts to prevent overfitting.
    \item Post-Hoc CBMs~\citep{yuksekgonul2022post}. A recent work that creates concept bottleneck layers then trains a linear layer along with a residual fitting layer for classification.
    \item Label-free CBMs~\citep{oikarinen2023label}: A recent work which uses concepts from GPT-3 then learns a projection matrix to approximate CLIP embeddings~\citep{radford2021learning}.
\end{itemize}

\paragraph{Implementation Details}
For all baselines, we use their public code bases. For fair comparisons, we replace the CLIP encoder in Post-Hoc CBM and Label-free CBM with the BioViL model, as a medical domain vision language model works better than CLIP.  
For more details regarding preprocessing, dataset statistics, baselines, and implementation,  see~\cref{app:implementation_details}.

\subsection{Robust Medical Image Classification}
\label{sec:exp_robust}
\paragraph{Results on datasets with explicit confounding factors} We first compare our method with baselines on the four curated datasets, as shown in~\cref{tab:main_robust}. 
\textbf{First}, standard methods to mitigate spurious correlations, such as Fish and LISA, may fail to capture the domain shift in this challenging setting. Moreover, they also require explicit confounding labels to better learn domain invariant features, which need careful expert analysis and are often difficult to obtain in real-world scenarios.
\textbf{Second}, image features can easily overfit to spurious correlations, even for a medical domain visual encoder such as BioViL. For example, it only has an accuracy of 9\% on the test set of NIH-age, meaning it learns to predict whether the chest x-ray is taken for a young or old individual almost perfectly, instead of learning to predict normal or abnormal.
\textbf{Finally}, our method also outperforms recent concept bottleneck models and attains much better robustness, demonstrating the effectiveness of our simple projection from visual features into a concept space.
\begin{table}[h!]
\vspace{-10pt}
    \caption{Performance comparison for robustness. Results are in percentage(\%).}
    \centering
    \resizebox{\linewidth}{!}{%
    \begin{tabular}{cccccc}
    \toprule
         Models & NIH-gender & NIH-age & NIH-agemix & Covid-mix & Interpretability\\
         \midrule
        ERM & 21.70 & 3.30 & 13.80 & 51.73 & \xmark\\
        Fish & 21.70 & 6.00 & 17.00 & 52.16 & \xmark\\
        LISA & 23.00 & 2.30 & 14.20 & 51.30 & \xmark \\
        \midrule
        BioViL Image Features & 71.60 & 9.40 & 13.70 & 51.08 & \xmark\\
        BioViL Image Features (dropouts) & 70.20 & 19.00 & 28.60 & 49.57 & \xmark\\
        \midrule
        Post-Hoc CBM & 77.40 & 13.70  & 16.70  & 51.08 & \checkmark\\
        Label-free CBM & 78.90 & 32.90  & 35.80 & 47.40 & \checkmark\\
        \midrule
        Ours & \textbf{79.60} & \textbf{50.70} & \textbf{53.40} & \textbf{62.36} & \checkmark\\ 
        \bottomrule
    \end{tabular}}
    \label{tab:main_robust}
    \vspace{-5pt}
\end{table}
\paragraph{Results on other benchmarks} We then evaluate the performance of our method on the other four datasets, which are popular benchmarks with no explicit confounding factors. As shown in~\cref{tab:main_standard}, on two datasets, projecting visual features into a concept space still leads to slightly better classification performance than latent visual features. We conjecture the reason could be the implicit confounding factors in the data, even though those datasets are assumed to be collected in an unbiased way where training and testing are drawn from the same distribution. Hence classification with concepts can improve model robustness and performance. 

\begin{table}[h!]
\vspace{-10pt}
\caption{Performance comparison on standard datasets without explicit confounding factors.}
    \centering
    \resizebox{\linewidth}{!}{%
    \begin{tabular}{cccccc}
    \toprule
         Models & NIH-CXR & Covid-QU & Pneumonia & Open-i & Interpretability\\
        \midrule
        BioViL Visual Encoder & \textbf{63.66} &  78.14 & 86.70 & 71.01 & \xmark\\
        BioViL Visual Encoder (dropouts) & 43.59 & 68.59 & 77.08 & 55.39 & \xmark\\
        \midrule
        Post-Hoc CBM & 62.88 & \textbf{79.09} & 87.34  &  72.35 & \checkmark\\
        Label-free CBM & 62.40 & 72.23  & 88.30 &  71.91 & \checkmark\\
        \midrule
        Ours &  63.27 & 78.00 & \textbf{88.46} & \textbf{72.80} & \checkmark\\ 
        \bottomrule
    \end{tabular}}
    \vspace{-10pt}
\label{tab:main_standard}
\end{table}

\subsection{Interpretable Medical Image Classification}
\label{sec:exp_interpretable}
%

Our framework offers three levels
of interpretation.
First, given a chest X-ray, the concept scores indicate the correlations between concepts and the chest X-ray. 
Second, as explained in~\cref{sec:method_fc}, our linear classification layer also offers interpretability; the model decisions can be explained as a simple linear combination of concepts. 
Lastly, the multiplications of concept scores and linear layer provide a way to interpret individual predictions. 

\begin{figure}[h!]
    \centering
    \includegraphics[width=0.49\linewidth]{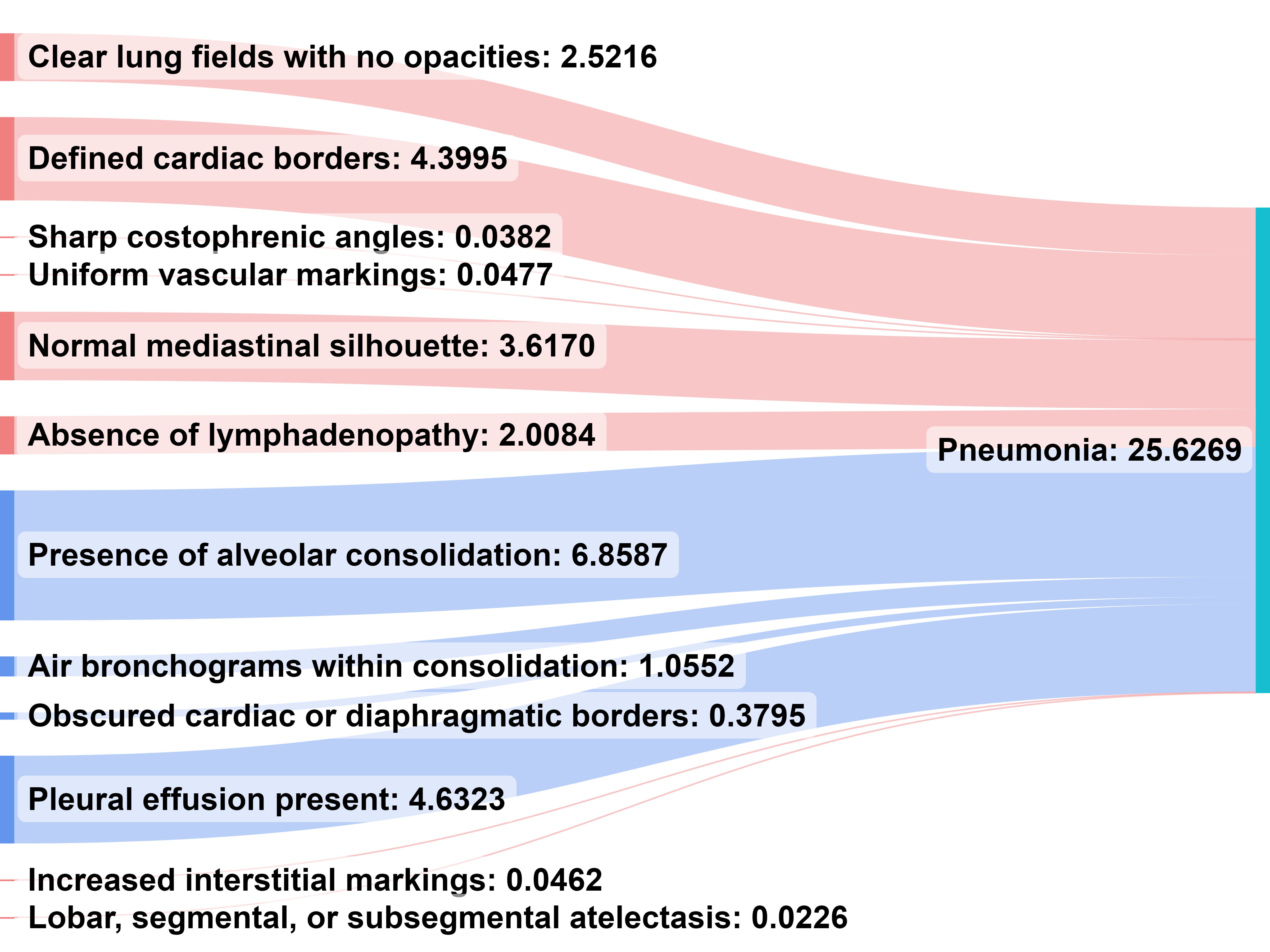}
    \hfill
    \includegraphics[width=0.49\linewidth]{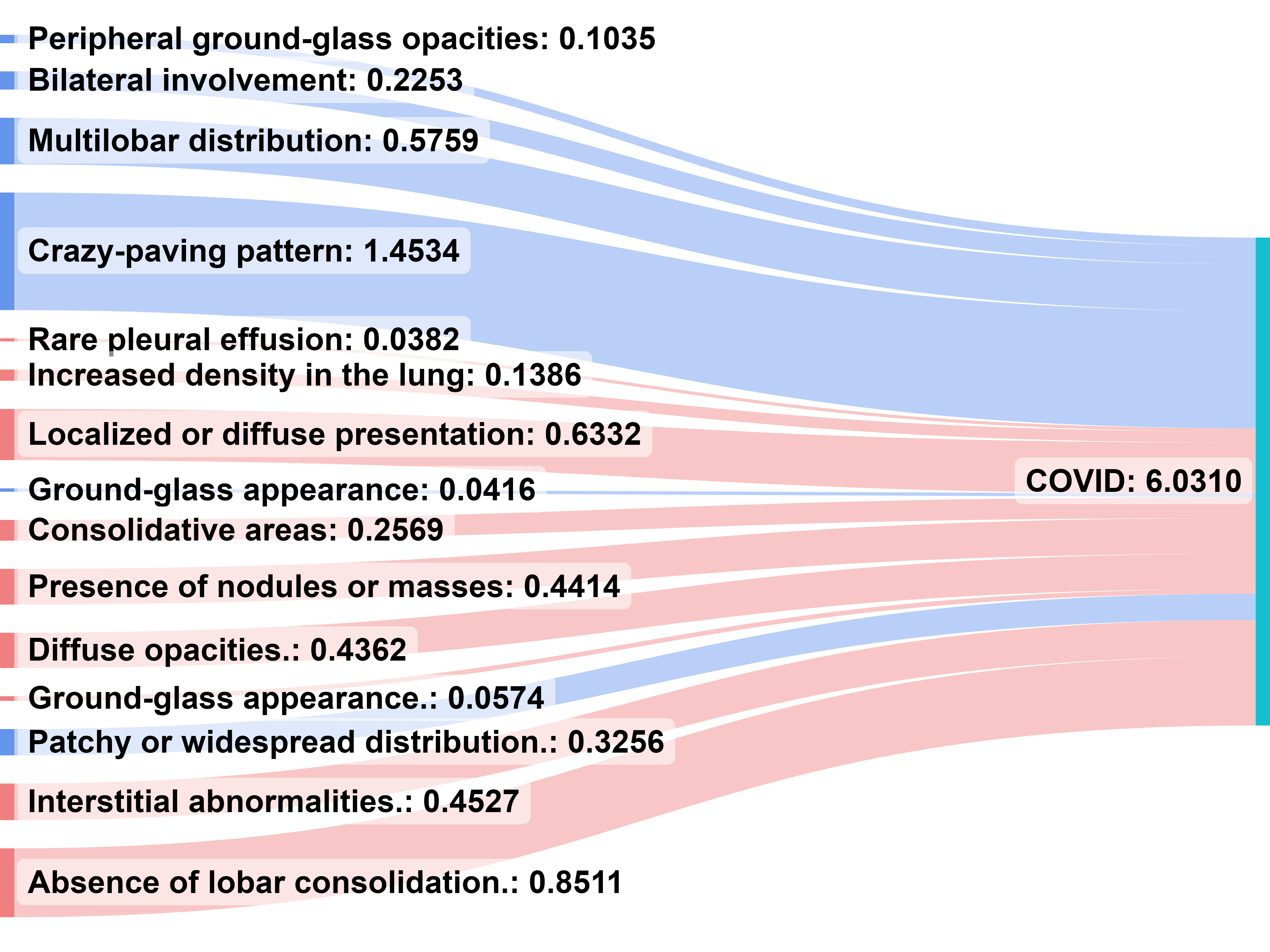}
    \caption{Visualization of the final layer weights on the pneumonia dataset and COVID-QU dataset. In each figure, the left side shows the concepts, while the right side shows the class. (best viewed in color, blue flows are positive weights and red flows are negative weights.)}
    \vspace{-10pt}
    \label{fig:pneumonia_gpt4_weights}
\end{figure}

\paragraph{Interpretation of learned linear weights.} We provide an example visualization of the learned final layer weights. For better visualization, we add L1 normalization to obtain sparse weights. We provide a Sankey diagram of the final layer weights on the two classes for the Pneumonia dataset shown in~\cref{fig:pneumonia_gpt4_weights}, where larger width of a line indicates larger weight between a concept and a class. 
As the prediction is a linear combination of concepts, we can interpret concepts with larger weights as more important to the class.
The visualization largely aligns with medical knowledge. For classifying pneumonia, the model treat pneumonia concepts as positive and normal concepts as negative. More interestingly, for classifying Covid-19, the model is able to find \textit{crazy-paving pattern} and \textit{multilobar distribution} as important positive concepts, which are identified by radiologists as two differential patterns in Covid-19 chest X-rays~\citep{chung2020ct,sheikhi2020review,salehi2020coronavirus}.

\begin{figure*}[h!]
\subfigure[ChestXray Pneumonia Instances]
{\label{fig:chestxray_pneumonia_instance}\includegraphics[width=0.475\linewidth]{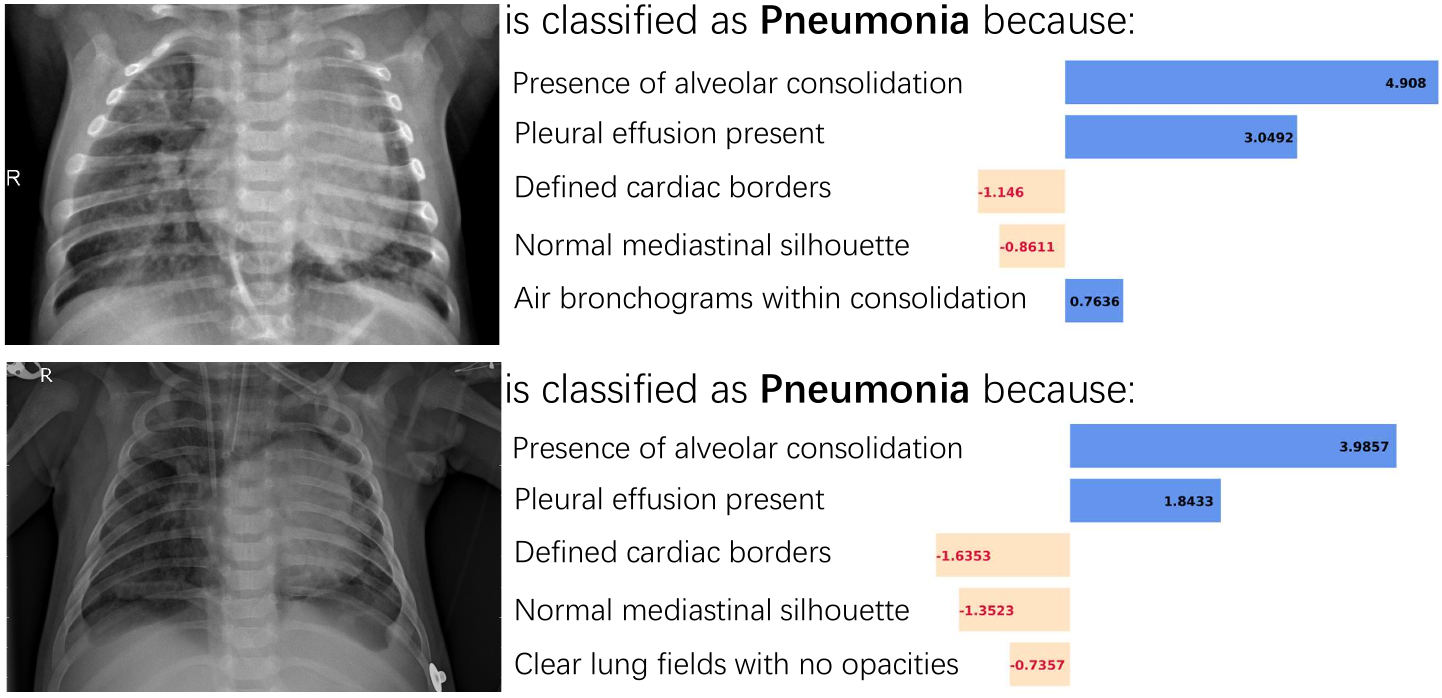}}
\subfigure[ChestXray Normal Instances]{\label{fig:chestxray_normal_instance}\includegraphics[width=0.475\linewidth]{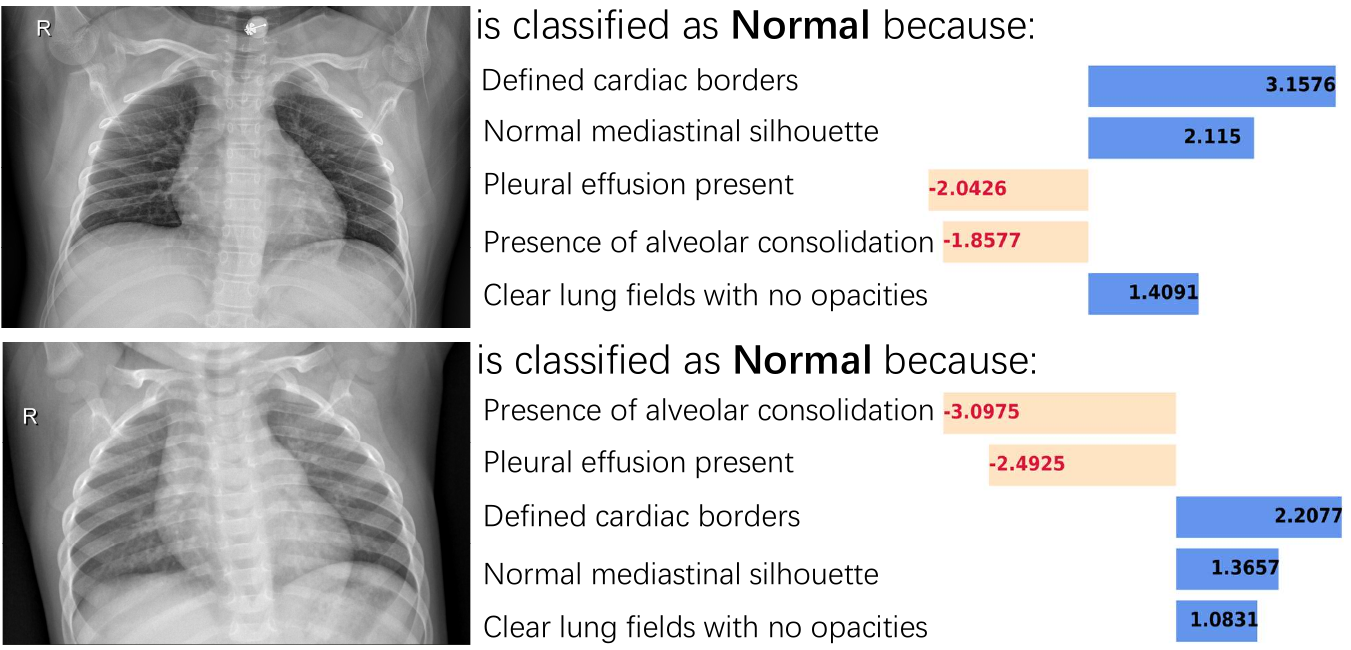}}
\caption{Examples of interpretability on the Pneumonia dataset}
\label{fig:chestxray_instance_interpretability}
\end{figure*}

\paragraph{Interpretation of instance predictions.} 
In addition to global understanding from the linear layer, our model can also interpret predictions of an instance. As shown in~\cref{fig:chestxray_instance_interpretability}, since the contribution of a concept $j$ to class $c$ on input image $I^{(i)}$ can be measured by $\textbf{W}^F_{j,c} s_j^{(i)}$, we visualize the interpretation of individual predictions with bar plots. For each Chest X-ray, concepts with top contribution scores are presented. Take the top right image as an example, \textit{Defined cardiac borders}, \textit{Normal mediastinal silhouette} and \textit{No Pleural effusion present} (negative weights indicate negations) are the important concepts to predict normal or pnuemonia for this X-ray. 

We provide more visualizations, case studies and human evaluation in~\cref{app:case_study}.


\subsection{Ablation Study}
\label{sec:exp_ablation}
\begin{figure}[h!]
    \centering
    \includegraphics[width=0.49\linewidth]{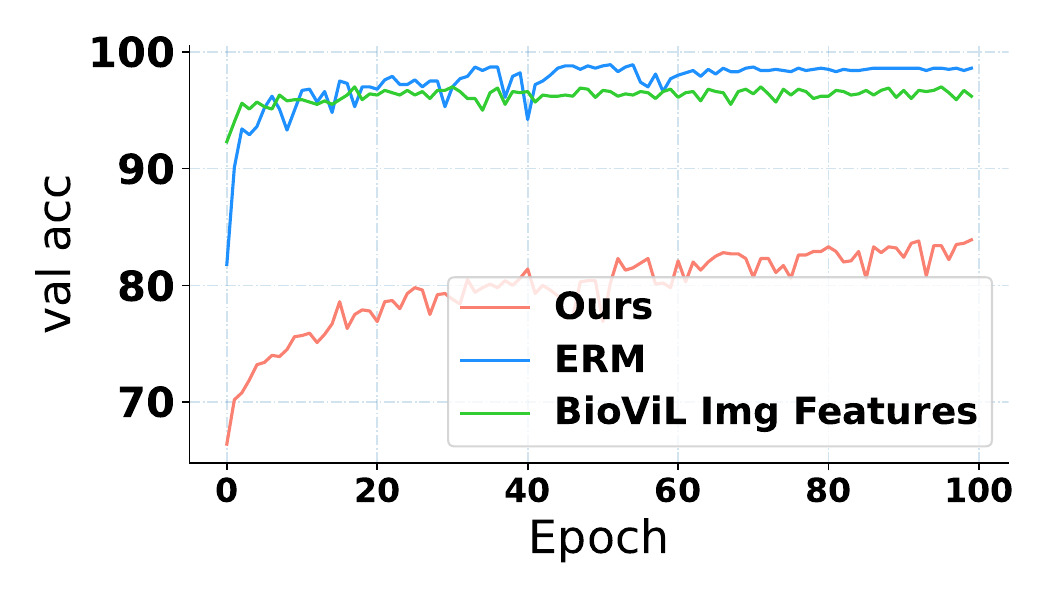}
    \includegraphics[width=0.49\linewidth]{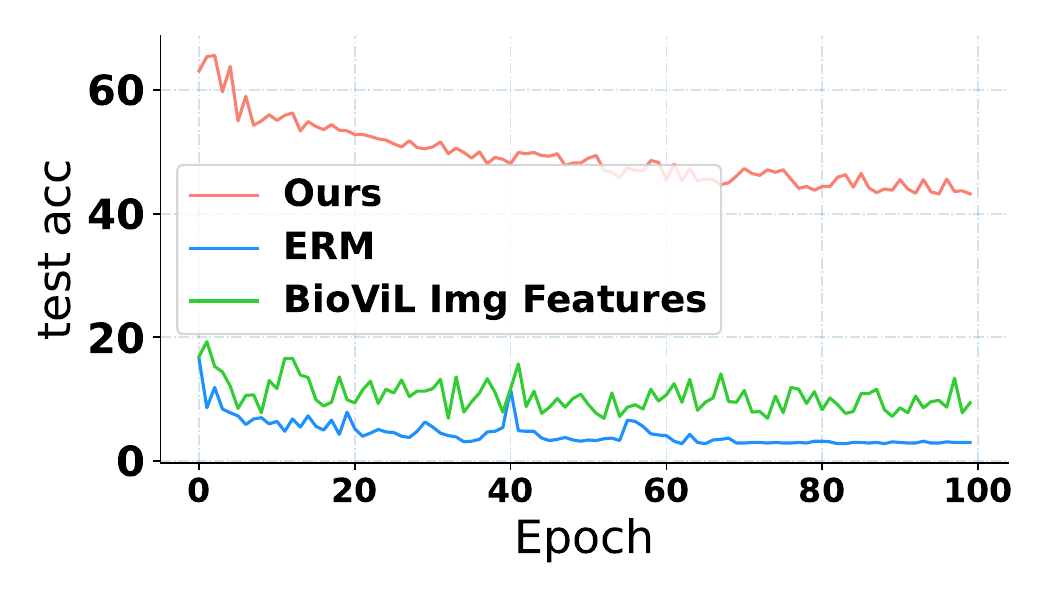}
    \caption{Validation and Test Accuracy along the training process on NIH-age }
    \label{fig:val_test_acc_wrt_epoch_nih_age}
\end{figure}

\paragraph{Training Curves}
To better understand the results on robustness, we visualize the validation and testing accuracy on NIH-age in~\cref{fig:val_test_acc_wrt_epoch_nih_age}. ERM and BioViL Image Features struggle to learn the critical information for classification, but instead learning shortcuts of the data, i.e., predicting if the X-ray is from a young or old individual near perfectly, hence achieving high validation accuracy but low test accuracy. In comparison, our method is much more robust to spurious correlations, with the accuracy being stable along the training process.

\paragraph{Comparing Different Concept Sets}
We further explore the effect of different concept sets. Results are presented in~\cref{tab:different_concept_sets}. We consider the following variations to compare with: 
(1) ChatGPT: Generating concepts using ChatGPT. 
(2) GPT4+Negation: Generating negations for all the concepts using GPT-4
(3) MIMIC-GPT4: Ask GPT-4 to read 100 reports from MIMIC-CXR~\citep{johnson2019mimic} and extract relevant concepts for each diseases. 
(4) Human: Descriptions annotated by radiologists in~\citep{boecking2022making}. We randomly select 12-28 descriptions to match the number of concepts from GPT-4.
Overall, we find GPT4 concepts perform the best in terms of accuracy. We also asked a board-certified radiologist to compare the concepts, and GPT-4 concepts are favored in terms of style and broad coverage.
\begin{table}[h!]
    \caption{Performance comparison for different concept sets.}
    \centering
    \begin{tabular}{cccccc}
    \toprule
         Models & NIH-gender & NIH-age & NIH-agemix & NIHCXR & Pneumonia \\
         \midrule
        ChatGPT & 79.40 & 44.90 & 48.40 & 63.18 & 88.30\\
        GPT4 & 79.60 & 50.70 & 53.40 & 63.27 & 88.46\\
        GPT4 + negation & 79.30 & 41.30 & 46.20 & 62.88 & 89.10\\
        MIMIC-GPT4 & 78.70 & 38.50 & 44.20 & 62.71 & 87.98\\
        Human & 79.20 & 44.80 & 47.70 & 63.06 & 88.62\\
        \bottomrule
    \end{tabular}
    \label{tab:different_concept_sets}
\end{table}

\begin{figure*}[h!]
\subfigure[NIH-gender]{\label{fig:nih-gender_number_of_concepts}\includegraphics[width=0.33\linewidth]{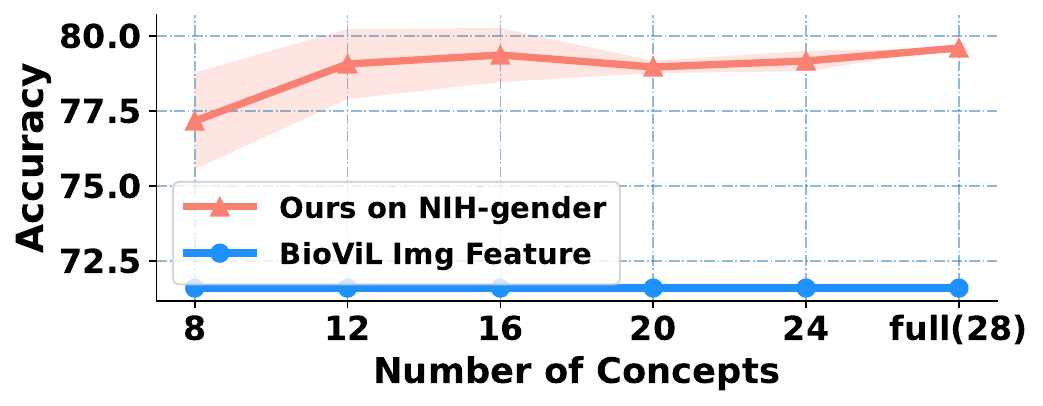}}
\subfigure[NIHCXR]{\label{fig:nihcxr_number_of_concepts}\includegraphics[width=0.33\linewidth]{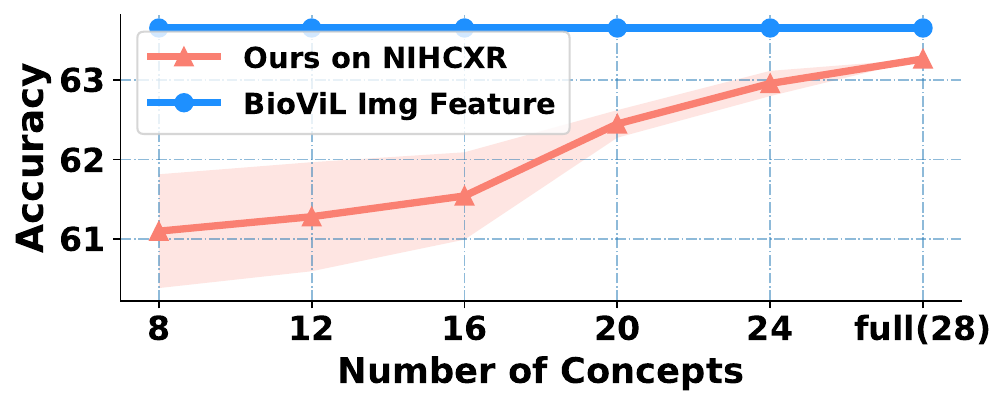}}
\subfigure[Pneumonia]{\label{fig:pneumonia_number_of_concepts}\includegraphics[width=0.33\linewidth]{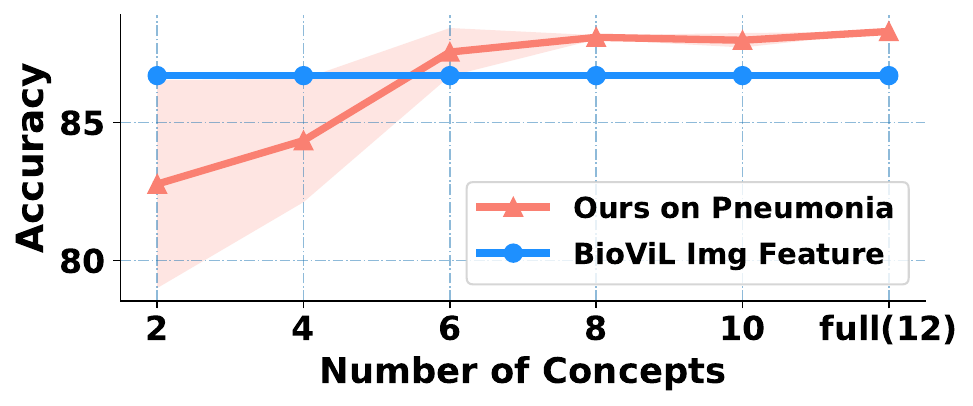}}
\caption{Comparison with different numbers of concepts $K$.}
\vspace{-2pt}
\label{fig:number_of_concepts}
\end{figure*}

\paragraph{Different number of concepts}
As shown in~\cref{fig:number_of_concepts}, we present classification results with different numbers of concepts, by randomly selecting $K$ concepts given a concept set. The results are mostly robust to the number of concepts. In general, a few number of concepts per class are necessary to maintain reasonable performance. 
Note that our concept vectors can achieve good classification with a much smaller number of dimensions than image features (e.g., 128).



\section{Related work}
\paragraph{Robustness and Spurious Correlations}
The robustness of machine learning models has been long studied, and the domain shift problem has drawn massive attention in past years~\citep{wilds}. To deal with this problem, some methods propose to learn the invariant representations~\citep{zhou2020domain,zhao2020maximum,
qiao2020learning,
volpi2018generalizing}, others aim at optimizing the worst-group accuracy when splitting the dataset into different groups based on the domains and labels~\citep{groupDRO,zhang2020coping,zhou2021examining,lisa}.  Most of these methods are limited by the requirement of domain labels, which are hard to obtain in real-world applications.
Our work differs in that we propose a simple yet effective method to remove spurious correlations for medical image classification, by projecting image features into a concept space with a pretrained vision-language model. 

\paragraph{Interpretable Image Classification}
Interpretability is important for healthcare applications with high-stakes 
decisions~\citep{yan2021weakly, rudin2022black,he2023nothing}. Various approaches have been proposed for interpretable medical models~\citep{zhang2017mdnet,sun2020saunet,barnett2021case,clough2019global,singh2020explainable,shen2021interpretable}. 
A line of research studies Concept Bottleneck Models (CBMs)~\citep{koh2020concept,chauhan2022interactive,zarlenga2022concept,yan2023learning}, which create a layer before the last fully connected layer where each neuron corresponds to a human interpretable concept. The initial CBMs require extensive human annotation for each image which is impractical to scale for many datasets. Recent work~\citep{yuksekgonul2022post,yang2022language,oikarinen2023label} proposes to connect visual features with concepts using vision-language models such as CLIP~\citep{radford2021learning}. 
However, to achieve competitive performance with image features, they either require learning additional residual connections~\citep{yuksekgonul2022post}, learning a projection matrix~\citep{oikarinen2023label}, or using a massive number of concepts~\citep{yang2022language}. 
In this paper, we show that those procedures are unnecessary and can even be problematic for learning robust representations for medical data. 
Instead, we present a minimalist design to achieve good performance on various benchmarks under different settings.

\paragraph{Large Language Models}
Large Language Models (LLMs) is an emergent topic in machine learning. 
By pretraining with a large amount of web data, they are shown to be adaptable to a wide range of downstream tasks in a few-shot or zero-shot manner~\citep{brown2020language,yan2022radbert,touvron2023llama}. 
Recent instruction-following LLMs~\citep{ouyang2022training}, e.g.,~ChatGPT and GPT-4~\citep{openai2023gpt}, provides an interface for human to interact with, by directly asking questions to the model.
In this paper, we leverage GPT-4 as a powerful and flexible tool to help with clinical natural language processing (e.g., querying concepts, extracting keywords from reports, revising descriptions), demonstrating its potential to assist in the development of healthcare applications.

\section{Limitations and Social Impact}
Our method requires access to a large language model, GPT-4. Querying ideal concepts that are aligned with human often needs a few rounds of revision of the prompts and interaction with the model. 
The generation results of GPT-4 come with randomness and are not perfectly controllable. 
Though the concepts are verified with a board-certified radiologist, there still could be biases induced by the model.
Successful estimations of the concept scores rely on a pretrained medical domain vision-language model, as inaccurate estimations will lead to wrong predictions. We present more details and analysis in~\cref{app:prompting} and~\cref{app:case_study}.

\section{Conclusion}
In this paper, we propose a new paradigm for building robust and interpretable medical image classifiers. By leveraging the concepts generated by LLMs, our model is able to mitigate spurious correlations and achieve strong results on the challenging datasets with explicit confounding factors, while also presenting a level of interpretability. 
Moreover, on standard datasets, it can still attain similar or even better performance than visual backbones, showcasing its potential to serve as a general paradigm to develop image classification models for healthcare. 



\clearpage

\small
\bibliographystyle{plainnat}
\bibliography{ref}

\begin{thebibliography}{61}
\providecommand{\natexlab}[1]{#1}
\providecommand{\url}[1]{\texttt{#1}}
\expandafter\ifx\csname urlstyle\endcsname\relax
  \providecommand{\doi}[1]{doi: #1}\else
  \providecommand{\doi}{doi: \begingroup \urlstyle{rm}\Url}\fi

\bibitem[Azizi et~al.(2021)Azizi, Mustafa, Ryan, Beaver, Freyberg, Deaton, Loh,
  Karthikesalingam, Kornblith, Chen, et~al.]{azizi2021big}
Shekoofeh Azizi, Basil Mustafa, Fiona Ryan, Zachary Beaver, Jan Freyberg,
  Jonathan Deaton, Aaron Loh, Alan Karthikesalingam, Simon Kornblith, Ting
  Chen, et~al.
\newblock Big self-supervised models advance medical image classification.
\newblock In \emph{Proceedings of the IEEE/CVF International Conference on
  Computer Vision}, pages 3478--3488, 2021.

\bibitem[Barnett et~al.(2021)Barnett, Schwartz, Tao, Chen, Ren, Lo, and
  Rudin]{barnett2021case}
Alina~Jade Barnett, Fides~Regina Schwartz, Chaofan Tao, Chaofan Chen, Yinhao
  Ren, Joseph~Y Lo, and Cynthia Rudin.
\newblock A case-based interpretable deep learning model for classification of
  mass lesions in digital mammography.
\newblock \emph{Nature Machine Intelligence}, 3\penalty0 (12):\penalty0
  1061--1070, 2021.

\bibitem[Boecking et~al.(2022)Boecking, Usuyama, Bannur, Castro, Schwaighofer,
  Hyland, Wetscherek, Naumann, Nori, Alvarez-Valle, et~al.]{boecking2022making}
Benedikt Boecking, Naoto Usuyama, Shruthi Bannur, Daniel~C Castro, Anton
  Schwaighofer, Stephanie Hyland, Maria Wetscherek, Tristan Naumann, Aditya
  Nori, Javier Alvarez-Valle, et~al.
\newblock Making the most of text semantics to improve biomedical
  vision--language processing.
\newblock In \emph{Computer Vision--ECCV 2022: 17th European Conference, Tel
  Aviv, Israel, October 23--27, 2022, Proceedings, Part XXXVI}, pages 1--21.
  Springer, 2022.

\bibitem[Brown et~al.(2020)Brown, Mann, Ryder, Subbiah, Kaplan, Dhariwal,
  Neelakantan, Shyam, Sastry, Askell, et~al.]{brown2020language}
Tom Brown, Benjamin Mann, Nick Ryder, Melanie Subbiah, Jared~D Kaplan, Prafulla
  Dhariwal, Arvind Neelakantan, Pranav Shyam, Girish Sastry, Amanda Askell,
  et~al.
\newblock Language models are few-shot learners.
\newblock \emph{Advances in neural information processing systems},
  33:\penalty0 1877--1901, 2020.

\bibitem[Chauhan et~al.(2022)Chauhan, Tiwari, Freyberg, Shenoy, and
  Dvijotham]{chauhan2022interactive}
Kushal Chauhan, Rishabh Tiwari, Jan Freyberg, Pradeep Shenoy, and Krishnamurthy
  Dvijotham.
\newblock Interactive concept bottleneck models.
\newblock \emph{arXiv preprint arXiv:2212.07430}, 2022.

\bibitem[Chowdhury et~al.(2020)Chowdhury, Rahman, Khandakar, Mazhar, Kadir,
  Mahbub, Islam, Khan, Iqbal, Al~Emadi, et~al.]{chowdhury2020can}
Muhammad~EH Chowdhury, Tawsifur Rahman, Amith Khandakar, Rashid Mazhar,
  Muhammad~Abdul Kadir, Zaid~Bin Mahbub, Khandakar~Reajul Islam,
  Muhammad~Salman Khan, Atif Iqbal, Nasser Al~Emadi, et~al.
\newblock Can ai help in screening viral and covid-19 pneumonia?
\newblock \emph{Ieee Access}, 8:\penalty0 132665--132676, 2020.

\bibitem[Chung et~al.(2020)Chung, Bernheim, Mei, Zhang, Huang, Zeng, Cui, Xu,
  Yang, Fayad, et~al.]{chung2020ct}
Michael Chung, Adam Bernheim, Xueyan Mei, Ning Zhang, Mingqian Huang, Xianjun
  Zeng, Jiufa Cui, Wenjian Xu, Yang Yang, Zahi~A Fayad, et~al.
\newblock Ct imaging features of 2019 novel coronavirus (2019-ncov).
\newblock \emph{Radiology}, 295\penalty0 (1):\penalty0 202--207, 2020.

\bibitem[Clough et~al.(2019)Clough, Oksuz, Puyol-Ant{\'o}n, Ruijsink, King, and
  Schnabel]{clough2019global}
James~R Clough, Ilkay Oksuz, Esther Puyol-Ant{\'o}n, Bram Ruijsink, Andrew~P
  King, and Julia~A Schnabel.
\newblock Global and local interpretability for cardiac mri classification.
\newblock In \emph{Medical Image Computing and Computer Assisted
  Intervention--MICCAI 2019: 22nd International Conference, Shenzhen, China,
  October 13--17, 2019, Proceedings, Part IV 22}, pages 656--664. Springer,
  2019.

\bibitem[Cohen et~al.(2020)Cohen, Morrison, and Dao]{cohen2020covid}
Joseph~Paul Cohen, Paul Morrison, and Lan Dao.
\newblock Covid-19 image data collection.
\newblock \emph{arXiv preprint arXiv:2003.11597}, 2020.

\bibitem[De~Bruijne(2016)]{de2016machine}
Marleen De~Bruijne.
\newblock Machine learning approaches in medical image analysis: From detection
  to diagnosis, 2016.

\bibitem[Demner-Fushman et~al.(2012)Demner-Fushman, Antani, Simpson, and
  Thoma]{demner2012design}
Dina Demner-Fushman, Sameer Antani, Matthew Simpson, and George~R Thoma.
\newblock Design and development of a multimodal biomedical information
  retrieval system.
\newblock \emph{Journal of Computing Science and Engineering}, 6\penalty0
  (2):\penalty0 168--177, 2012.

\bibitem[Harmon et~al.(2020)Harmon, Sanford, Xu, Turkbey, Roth, Xu, Yang,
  Myronenko, Anderson, Amalou, et~al.]{harmon2020artificial}
Stephanie~A Harmon, Thomas~H Sanford, Sheng Xu, Evrim~B Turkbey, Holger Roth,
  Ziyue Xu, Dong Yang, Andriy Myronenko, Victoria Anderson, Amel Amalou, et~al.
\newblock Artificial intelligence for the detection of covid-19 pneumonia on
  chest ct using multinational datasets.
\newblock \emph{Nature communications}, 11\penalty0 (1):\penalty0 4080, 2020.

\bibitem[He et~al.(2023)He, Yan, Gentili, McAuley, and Hsu]{he2023nothing}
Zexue He, An~Yan, Amilcare Gentili, Julian McAuley, and Chun-Nan Hsu.
\newblock " nothing abnormal": Disambiguating medical reports via contrastive
  knowledge infusion.
\newblock \emph{arXiv preprint arXiv:2305.08300}, 2023.

\bibitem[Johnson et~al.(2019)Johnson, Pollard, Berkowitz, Greenbaum, Lungren,
  Deng, Mark, and Horng]{johnson2019mimic}
Alistair~EW Johnson, Tom~J Pollard, Seth~J Berkowitz, Nathaniel~R Greenbaum,
  Matthew~P Lungren, Chih-ying Deng, Roger~G Mark, and Steven Horng.
\newblock Mimic-cxr, a de-identified publicly available database of chest
  radiographs with free-text reports.
\newblock \emph{Scientific data}, 6\penalty0 (1):\penalty0 317, 2019.

\bibitem[Kermany et~al.(2018)Kermany, Goldbaum, Cai, Valentim, Liang, Baxter,
  McKeown, Yang, Wu, Yan, et~al.]{kermany2018identifying}
Daniel~S Kermany, Michael Goldbaum, Wenjia Cai, Carolina~CS Valentim, Huiying
  Liang, Sally~L Baxter, Alex McKeown, Ge~Yang, Xiaokang Wu, Fangbing Yan,
  et~al.
\newblock Identifying medical diagnoses and treatable diseases by image-based
  deep learning.
\newblock \emph{cell}, 172\penalty0 (5):\penalty0 1122--1131, 2018.

\bibitem[Koh et~al.(2020)Koh, Nguyen, Tang, Mussmann, Pierson, Kim, and
  Liang]{koh2020concept}
Pang~Wei Koh, Thao Nguyen, Yew~Siang Tang, Stephen Mussmann, Emma Pierson, Been
  Kim, and Percy Liang.
\newblock Concept bottleneck models.
\newblock In \emph{International Conference on Machine Learning}, pages
  5338--5348. PMLR, 2020.

\bibitem[Koh et~al.(2021)Koh, Sagawa, Marklund, Xie, Zhang, Balsubramani, Hu,
  Yasunaga, Phillips, Gao, Lee, David, Stavness, Guo, Earnshaw, Haque, Beery,
  Leskovec, Kundaje, Pierson, Levine, Finn, and Liang]{wilds}
Pang~Wei Koh, Shiori Sagawa, Henrik Marklund, Sang~Michael Xie, Marvin Zhang,
  Akshay Balsubramani, Weihua Hu, Michihiro Yasunaga, Richard~Lanas Phillips,
  Irena Gao, Tony Lee, Etienne David, Ian Stavness, Wei Guo, Berton Earnshaw,
  Imran~S. Haque, Sara~M. Beery, Jure Leskovec, Anshul Kundaje, Emma Pierson,
  Sergey Levine, Chelsea Finn, and Percy Liang.
\newblock {WILDS:} {A} benchmark of in-the-wild distribution shifts.
\newblock In \emph{{ICML}}, volume 139 of \emph{Proceedings of Machine Learning
  Research}, pages 5637--5664. {PMLR}, 2021.

\bibitem[Kojima et~al.(2022)Kojima, Gu, Reid, Matsuo, and
  Iwasawa]{kojima2022large}
Takeshi Kojima, Shixiang~Shane Gu, Machel Reid, Yutaka Matsuo, and Yusuke
  Iwasawa.
\newblock Large language models are zero-shot reasoners.
\newblock \emph{arXiv preprint arXiv:2205.11916}, 2022.

\bibitem[Litjens et~al.(2017)Litjens, Kooi, Bejnordi, Setio, Ciompi,
  Ghafoorian, Van Der~Laak, Van~Ginneken, and S{\'a}nchez]{litjens2017survey}
Geert Litjens, Thijs Kooi, Babak~Ehteshami Bejnordi, Arnaud Arindra~Adiyoso
  Setio, Francesco Ciompi, Mohsen Ghafoorian, Jeroen~Awm Van Der~Laak, Bram
  Van~Ginneken, and Clara~I S{\'a}nchez.
\newblock A survey on deep learning in medical image analysis.
\newblock \emph{Medical image analysis}, 42:\penalty0 60--88, 2017.

\bibitem[Liu et~al.(2018)Liu, Luo, Wang, and Tang]{liu2018large}
Ziwei Liu, Ping Luo, Xiaogang Wang, and Xiaoou Tang.
\newblock Large-scale celebfaces attributes (celeba) dataset.
\newblock \emph{Retrieved August}, 15\penalty0 (2018):\penalty0 11, 2018.

\bibitem[Lu et~al.(2022)Lu, Yan, Chang, Hsu, McAuley, Du, and
  Gentili]{lu2022semi}
Xing Lu, An~Yan, Eric~Y Chang, C-n Hsu, Julian McAuley, Jiang Du, and Amilcare
  Gentili.
\newblock Semi-supervised multi-label classification with 3d cbam resnet for
  tuberculosis cavern report.
\newblock In \emph{CLEF2022 Working Notes, CEUR Workshop Proceedings, CEUR-WS.
  org< http://ceurws. org>, Bologna, Italy}, 2022.

\bibitem[Menon and Vondrick(2022)]{menon2022visual}
Sachit Menon and Carl Vondrick.
\newblock Visual classification via description from large language models.
\newblock \emph{arXiv preprint arXiv:2210.07183}, 2022.

\bibitem[Nori et~al.(2023)Nori, King, McKinney, Carignan, and
  Horvitz]{nori2023capabilities}
Harsha Nori, Nicholas King, Scott~Mayer McKinney, Dean Carignan, and Eric
  Horvitz.
\newblock Capabilities of gpt-4 on medical challenge problems.
\newblock \emph{arXiv preprint arXiv:2303.13375}, 2023.

\bibitem[Oikarinen et~al.(2023)Oikarinen, Das, Nguyen, and
  Weng]{oikarinen2023label}
Tuomas Oikarinen, Subhro Das, Lam~M Nguyen, and Tsui-Wei Weng.
\newblock Label-free concept bottleneck models.
\newblock \emph{arXiv preprint arXiv:2304.06129}, 2023.

\bibitem[OpenAI(2023)]{openai2023gpt}
OpenAI.
\newblock Gpt-4 technical report.
\newblock \emph{arXiv}, 2023.

\bibitem[Ouyang et~al.(2022)Ouyang, Wu, Jiang, Almeida, Wainwright, Mishkin,
  Zhang, Agarwal, Slama, Ray, et~al.]{ouyang2022training}
Long Ouyang, Jeffrey Wu, Xu~Jiang, Diogo Almeida, Carroll Wainwright, Pamela
  Mishkin, Chong Zhang, Sandhini Agarwal, Katarina Slama, Alex Ray, et~al.
\newblock Training language models to follow instructions with human feedback.
\newblock \emph{Advances in Neural Information Processing Systems},
  35:\penalty0 27730--27744, 2022.

\bibitem[Qiao et~al.(2020)Qiao, Zhao, and Peng]{qiao2020learning}
Fengchun Qiao, Long Zhao, and Xi~Peng.
\newblock Learning to learn single domain generalization.
\newblock In \emph{Proceedings of the IEEE/CVF Conference on Computer Vision
  and Pattern Recognition}, pages 12556--12565, 2020.

\bibitem[Radford et~al.(2021{\natexlab{a}})Radford, Kim, Hallacy, Ramesh, Goh,
  Agarwal, Sastry, Askell, Mishkin, Clark, Krueger, and Sutskever]{CLIP}
Alec Radford, Jong~Wook Kim, Chris Hallacy, Aditya Ramesh, Gabriel Goh,
  Sandhini Agarwal, Girish Sastry, Amanda Askell, Pamela Mishkin, Jack Clark,
  Gretchen Krueger, and Ilya Sutskever.
\newblock Learning transferable visual models from natural language
  supervision.
\newblock In \emph{{ICML}}, volume 139 of \emph{Proceedings of Machine Learning
  Research}, pages 8748--8763. {PMLR}, 2021{\natexlab{a}}.

\bibitem[Radford et~al.(2021{\natexlab{b}})Radford, Kim, Hallacy, Ramesh, Goh,
  Agarwal, Sastry, Askell, Mishkin, Clark, et~al.]{radford2021learning}
Alec Radford, Jong~Wook Kim, Chris Hallacy, Aditya Ramesh, Gabriel Goh,
  Sandhini Agarwal, Girish Sastry, Amanda Askell, Pamela Mishkin, Jack Clark,
  et~al.
\newblock Learning transferable visual models from natural language
  supervision.
\newblock In \emph{International conference on machine learning}, pages
  8748--8763. PMLR, 2021{\natexlab{b}}.

\bibitem[Rudin(2022)]{rudin2022black}
Cynthia Rudin.
\newblock Why black box machine learning should be avoided for high-stakes
  decisions, in brief.
\newblock \emph{Nature Reviews Methods Primers}, 2\penalty0 (1):\penalty0 81,
  2022.

\bibitem[Sagawa et~al.(2020{\natexlab{a}})Sagawa, Koh, Hashimoto, and
  Liang]{groupDRO}
Shiori Sagawa, Pang~Wei Koh, Tatsunori~B Hashimoto, and Percy Liang.
\newblock Distributionally robust neural networks for group shifts: On the
  importance of regularization for worst-case generalization.
\newblock In \emph{ICLR}, 2020{\natexlab{a}}.

\bibitem[Sagawa et~al.(2020{\natexlab{b}})Sagawa, Raghunathan, Koh, and
  Liang]{sagawa2020investigation}
Shiori Sagawa, Aditi Raghunathan, Pang~Wei Koh, and Percy Liang.
\newblock An investigation of why overparameterization exacerbates spurious
  correlations.
\newblock In \emph{International Conference on Machine Learning}, pages
  8346--8356. PMLR, 2020{\natexlab{b}}.

\bibitem[Salehi et~al.(2020)Salehi, Abedi, Balakrishnan, Gholamrezanezhad,
  et~al.]{salehi2020coronavirus}
Sana Salehi, Aidin Abedi, Sudheer Balakrishnan, Ali Gholamrezanezhad, et~al.
\newblock Coronavirus disease 2019 (covid-19): a systematic review of imaging
  findings in 919 patients.
\newblock \emph{Ajr Am J Roentgenol}, 215\penalty0 (1):\penalty0 87--93, 2020.

\bibitem[Santa~Cruz et~al.(2021)Santa~Cruz, Bossa, S{\"o}lter, and
  Husch]{santa2021public}
Beatriz~Garcia Santa~Cruz, Mat{\'\i}as~Nicol{\'a}s Bossa, Jan S{\"o}lter, and
  Andreas~Dominik Husch.
\newblock Public covid-19 x-ray datasets and their impact on model bias--a
  systematic review of a significant problem.
\newblock \emph{Medical image analysis}, 74:\penalty0 102225, 2021.

\bibitem[Shamshad et~al.(2023)Shamshad, Khan, Zamir, Khan, Hayat, Khan, and
  Fu]{shamshad2023transformers}
Fahad Shamshad, Salman Khan, Syed~Waqas Zamir, Muhammad~Haris Khan, Munawar
  Hayat, Fahad~Shahbaz Khan, and Huazhu Fu.
\newblock Transformers in medical imaging: A survey.
\newblock \emph{Medical Image Analysis}, page 102802, 2023.

\bibitem[Sheikhi et~al.(2020)Sheikhi, Shirzadfar, and
  Sheikhi]{sheikhi2020review}
Kowsar Sheikhi, Hamidreza Shirzadfar, and Milad Sheikhi.
\newblock A review on novel coronavirus (covid-19): symptoms, transmission and
  diagnosis tests.
\newblock \emph{Research in Infectious Diseases and Tropical Medicine},
  2\penalty0 (1):\penalty0 1--8, 2020.

\bibitem[Shen et~al.(2021)Shen, Wu, Phang, Park, Liu, Tyagi, Heacock, Kim, Moy,
  Cho, et~al.]{shen2021interpretable}
Yiqiu Shen, Nan Wu, Jason Phang, Jungkyu Park, Kangning Liu, Sudarshini Tyagi,
  Laura Heacock, S~Gene Kim, Linda Moy, Kyunghyun Cho, et~al.
\newblock An interpretable classifier for high-resolution breast cancer
  screening images utilizing weakly supervised localization.
\newblock \emph{Medical image analysis}, 68:\penalty0 101908, 2021.

\bibitem[Shi et~al.(2021)Shi, Seely, Torr, Siddharth, Hannun, Usunier, and
  Synnaeve]{shi2021gradient}
Yuge Shi, Jeffrey Seely, Philip~HS Torr, N~Siddharth, Awni Hannun, Nicolas
  Usunier, and Gabriel Synnaeve.
\newblock Gradient matching for domain generalization.
\newblock \emph{arXiv preprint arXiv:2104.09937}, 2021.

\bibitem[Singh et~al.(2020)Singh, Sengupta, and
  Lakshminarayanan]{singh2020explainable}
Amitojdeep Singh, Sourya Sengupta, and Vasudevan Lakshminarayanan.
\newblock Explainable deep learning models in medical image analysis.
\newblock \emph{Journal of Imaging}, 6\penalty0 (6):\penalty0 52, 2020.

\bibitem[Sun et~al.(2020)Sun, Darbehani, Zaidi, and Wang]{sun2020saunet}
Jesse Sun, Fatemeh Darbehani, Mark Zaidi, and Bo~Wang.
\newblock Saunet: Shape attentive u-net for interpretable medical image
  segmentation.
\newblock In \emph{Medical Image Computing and Computer Assisted
  Intervention--MICCAI 2020: 23rd International Conference, Lima, Peru, October
  4--8, 2020, Proceedings, Part IV 23}, pages 797--806. Springer, 2020.

\bibitem[Thomas et~al.(2021)Thomas, Lefevre, Baxter, and
  Hamilton]{thomas2021interpretable}
Simon~M Thomas, James~G Lefevre, Glenn Baxter, and Nicholas~A Hamilton.
\newblock Interpretable deep learning systems for multi-class segmentation and
  classification of non-melanoma skin cancer.
\newblock \emph{Medical Image Analysis}, 68:\penalty0 101915, 2021.

\bibitem[Touvron et~al.(2023)Touvron, Lavril, Izacard, Martinet, Lachaux,
  Lacroix, Rozi{\`e}re, Goyal, Hambro, Azhar, et~al.]{touvron2023llama}
Hugo Touvron, Thibaut Lavril, Gautier Izacard, Xavier Martinet, Marie-Anne
  Lachaux, Timoth{\'e}e Lacroix, Baptiste Rozi{\`e}re, Naman Goyal, Eric
  Hambro, Faisal Azhar, et~al.
\newblock Llama: Open and efficient foundation language models.
\newblock \emph{arXiv preprint arXiv:2302.13971}, 2023.

\bibitem[Van~der Velden et~al.(2022)Van~der Velden, Kuijf, Gilhuijs, and
  Viergever]{van2022explainable}
Bas~HM Van~der Velden, Hugo~J Kuijf, Kenneth~GA Gilhuijs, and Max~A Viergever.
\newblock Explainable artificial intelligence (xai) in deep learning-based
  medical image analysis.
\newblock \emph{Medical Image Analysis}, page 102470, 2022.

\bibitem[Volpi et~al.(2018)Volpi, Namkoong, Sener, Duchi, Murino, and
  Savarese]{volpi2018generalizing}
Riccardo Volpi, Hongseok Namkoong, Ozan Sener, John Duchi, Vittorio Murino, and
  Silvio Savarese.
\newblock Generalizing to unseen domains via adversarial data augmentation.
\newblock \emph{arXiv preprint arXiv:1805.12018}, 2018.

\bibitem[Wang et~al.(2017)Wang, Peng, Lu, Lu, Bagheri, and
  Summers]{wang2017chestx}
Xiaosong Wang, Yifan Peng, Le~Lu, Zhiyong Lu, Mohammadhadi Bagheri, and
  Ronald~M Summers.
\newblock Chestx-ray8: Hospital-scale chest x-ray database and benchmarks on
  weakly-supervised classification and localization of common thorax diseases.
\newblock In \emph{Proceedings of the IEEE conference on computer vision and
  pattern recognition}, pages 2097--2106, 2017.

\bibitem[Wang et~al.(2022{\natexlab{a}})Wang, Ye, Zhu, Peng, Tian, and
  Zhu]{wang2022metateacher}
Zhenbin Wang, Mao Ye, Xiatian Zhu, Liuhan Peng, Liang Tian, and Yingying Zhu.
\newblock Metateacher: Coordinating multi-model domain adaptation for medical
  image classification.
\newblock \emph{Advances in Neural Information Processing Systems},
  35:\penalty0 20823--20837, 2022{\natexlab{a}}.

\bibitem[Wang et~al.(2022{\natexlab{b}})Wang, Wu, Agarwal, and Sun]{medclip}
Zifeng Wang, Zhenbang Wu, Dinesh Agarwal, and Jimeng Sun.
\newblock Medclip: Contrastive learning from unpaired medical images and text.
\newblock In \emph{{EMNLP}}, pages 3876--3887. Association for Computational
  Linguistics, 2022{\natexlab{b}}.

\bibitem[Wei et~al.(2022)Wei, Wang, Schuurmans, Bosma, Chi, Le, and
  Zhou]{wei2022chain}
Jason Wei, Xuezhi Wang, Dale Schuurmans, Maarten Bosma, Ed~Chi, Quoc Le, and
  Denny Zhou.
\newblock Chain of thought prompting elicits reasoning in large language
  models.
\newblock \emph{arXiv preprint arXiv:2201.11903}, 2022.

\bibitem[Yan et~al.(2021)Yan, He, Lu, Du, Chang, Gentili, McAuley, and
  Hsu]{yan2021weakly}
An~Yan, Zexue He, Xing Lu, Jiang Du, Eric Chang, Amilcare Gentili, Julian
  McAuley, and Chun-Nan Hsu.
\newblock Weakly supervised contrastive learning for chest x-ray report
  generation.
\newblock \emph{arXiv preprint arXiv:2109.12242}, 2021.

\bibitem[Yan et~al.(2022)Yan, McAuley, Lu, Du, Chang, Gentili, and
  Hsu]{yan2022radbert}
An~Yan, Julian McAuley, Xing Lu, Jiang Du, Eric~Y Chang, Amilcare Gentili, and
  Chun-Nan Hsu.
\newblock Radbert: Adapting transformer-based language models to radiology.
\newblock \emph{Radiology: Artificial Intelligence}, 4\penalty0 (4):\penalty0
  e210258, 2022.

\bibitem[Yan et~al.(2023)Yan, Wang, Zhong, Dong, He, Lu, Wang, Shang, and
  McAuley]{yan2023learning}
An~Yan, Yu~Wang, Yiwu Zhong, Chengyu Dong, Zexue He, Yujie Lu, William~Yang
  Wang, Jingbo Shang, and Julian McAuley.
\newblock Learning concise and descriptive attributes for visual recognition.
\newblock In \emph{Proceedings of the IEEE/CVF International Conference on
  Computer Vision}, pages 3090--3100, 2023.

\bibitem[Yang et~al.(2022)Yang, Panagopoulou, Zhou, Jin, Callison-Burch, and
  Yatskar]{yang2022language}
Yue Yang, Artemis Panagopoulou, Shenghao Zhou, Daniel Jin, Chris
  Callison-Burch, and Mark Yatskar.
\newblock Language in a bottle: Language model guided concept bottlenecks for
  interpretable image classification.
\newblock \emph{arXiv preprint arXiv:2211.11158}, 2022.

\bibitem[Yao et~al.(2022)Yao, Wang, Li, Zhang, Liang, Zou, and Finn]{lisa}
Huaxiu Yao, Yu~Wang, Sai Li, Linjun Zhang, Weixin Liang, James Zou, and Chelsea
  Finn.
\newblock Improving out-of-distribution robustness via selective augmentation.
\newblock In \emph{{ICML}}, volume 162 of \emph{Proceedings of Machine Learning
  Research}, pages 25407--25437. {PMLR}, 2022.

\bibitem[Yuksekgonul et~al.(2022)Yuksekgonul, Wang, and
  Zou]{yuksekgonul2022post}
Mert Yuksekgonul, Maggie Wang, and James Zou.
\newblock Post-hoc concept bottleneck models.
\newblock \emph{arXiv preprint arXiv:2205.15480}, 2022.

\bibitem[Zarlenga et~al.(2022)Zarlenga, Pietro, Gabriele, Giuseppe, Giannini,
  Diligenti, Zohreh, Frederic, Melacci, Adrian, et~al.]{zarlenga2022concept}
Mateo~Espinosa Zarlenga, Barbiero Pietro, Ciravegna Gabriele, Marra Giuseppe,
  Francesco Giannini, Michelangelo Diligenti, Shams Zohreh, Precioso Frederic,
  Stefano Melacci, Weller Adrian, et~al.
\newblock Concept embedding models: Beyond the accuracy-explainability
  trade-off.
\newblock In \emph{Advances in Neural Information Processing Systems},
  volume~35, pages 21400--21413. Curran Associates, Inc., 2022.

\bibitem[Zech et~al.(2018)Zech, Badgeley, Liu, Costa, Titano, and
  Oermann]{zech2018confounding}
John~R Zech, Marcus~A Badgeley, Manway Liu, Anthony~B Costa, Joseph~J Titano,
  and Eric~K Oermann.
\newblock Confounding variables can degrade generalization performance of
  radiological deep learning models.
\newblock \emph{arXiv preprint arXiv:1807.00431}, 2018.

\bibitem[Zhang et~al.(2021)Zhang, Menon, Veit, Bhojanapalli, Kumar, and
  Sra]{zhang2020coping}
Jingzhao Zhang, Aditya Menon, Andreas Veit, Srinadh Bhojanapalli, Sanjiv Kumar,
  and Suvrit Sra.
\newblock Coping with label shift via distributionally robust optimisation.
\newblock In \emph{ICLR}, 2021.

\bibitem[Zhang et~al.(2017)Zhang, Xie, Xing, McGough, and Yang]{zhang2017mdnet}
Zizhao Zhang, Yuanpu Xie, Fuyong Xing, Mason McGough, and Lin Yang.
\newblock Mdnet: A semantically and visually interpretable medical image
  diagnosis network.
\newblock In \emph{Proceedings of the IEEE conference on computer vision and
  pattern recognition}, pages 6428--6436, 2017.

\bibitem[Zhao et~al.(2020)Zhao, Liu, Peng, and Metaxas]{zhao2020maximum}
Long Zhao, Ting Liu, Xi~Peng, and Dimitris Metaxas.
\newblock Maximum-entropy adversarial data augmentation for improved
  generalization and robustness.
\newblock \emph{arXiv preprint arXiv:2010.08001}, 2020.

\bibitem[Zhou et~al.(2021)Zhou, Ma, Michel, and Neubig]{zhou2021examining}
Chunting Zhou, Xuezhe Ma, Paul Michel, and Graham Neubig.
\newblock Examining and combating spurious features under distribution shift.
\newblock In \emph{ICML}, 2021.

\bibitem[Zhou et~al.(2020)Zhou, Jiang, Shui, Wang, and
  Chaib-draa]{zhou2020domain}
Fan Zhou, Zhuqing Jiang, Changjian Shui, Boyu Wang, and Brahim Chaib-draa.
\newblock Domain generalization with optimal transport and metric learning.
\newblock \emph{arXiv preprint arXiv:2007.10573}, 2020.

\end{thebibliography}

\clearpage
\appendix
\section{Prompting Large Language Models}
\label{app:prompting}

\subsection{Interaction with GPT-4}
In this work, we leverage GPT-4 to generate useful radiology concepts.
As GPT-4 is a instruction-following model~\citep{ouyang2022training}, we can directly ask questions to interact with the model.

We start with the following conversation:\\
\textit{User: What are the useful visual attributes to recognize \{Atelectasis\} in a chest X-ray?\\GPT-4: Atelectasis refers to the partial or complete collapse of the lung or a portion of it. When it comes to recognizing atelectasis in a chest X-ray, there are several visual attributes that can be helpful. However, it's important to note that the definitive diagnosis of atelectasis requires the expertise of a qualified medical professional. That being said, here are some useful visual attributes that may indicate atelectasis in a chest X-ray:\\
1. Lung Opacity: Atelectasis typically presents as an area of increased density or opacity in the affected lung region. This may appear as a white or gray area on the X-ray, indicating reduced air content.\\
2. Volume Loss: ...}

The output of a initial prompt may be verbose and is not ideally what we want. We can add further conditions and constraints to obtain concise concepts:\\
\textit{User: Can you generate concise attributes for Atelectasis?\\
GPT-4:
Sure, here's a concise summary of the key attributes to identify Atelectasis on a chest X-ray:\\
Increased Opacity: Denser area in the lung field.\\
Volume Loss: Apparent shrinkage of affected lung region.\\
Displacement of Structures: Movement of thoracic structures toward the affected area.\\
Loss of Lung Markings: Absence of normal vascular and bronchial markings in the affected area.\\
Crowding of Ribs: Closer rib spacing on the side of the atelectasis.\\
Visible Bronchograms: Air-filled bronchi in the region of atelectasis.\\
Compensatory Overinflation: Adjacent or opposite lung fields may appear overinflated.
}

We find the following prompt works well:\\
\textit{User: Can you provide concise radiology descriptors for  Atelectasis? List in bullet points with no extra context.\\
GPT-4:
Sure, here are the concise radiology descriptors for Atelectasis:\\
Increased opacity\\
Lung volume loss\\
Displacement of thoracic structures\\
Loss of lung markings\\
Rib crowding\\
Visible bronchograms\\
Compensatory overinflation}

\subsection{Robustness Check}
To confirm the effectiveness of our prompts and the robustness of GPT-4 prompts, we conduct the experiments with the concepts queried from GPT-4 using different prompts. We design semantically instructive and misleading prompts as shown in~\cref{tab:robustness_check_prompts}. Overall, we observe that other instructive prompts perform similar to ours, while misleading prompts could hurt the performance drastically.

\begin{table}[h!]
    \centering
    \caption{Robustness study against different prompts on NIHCXR}
    \resizebox{\linewidth}{!}{%
    \begin{tabular}{c|p{14cm}}
        \toprule
 Category &  \begin{tabular}[c]{p{0.9\linewidth}|c} \begin{tabular}[c]{@{}c@{}} Prompts \end{tabular} & Acc \end{tabular} \\
 \midrule
 \begin{tabular}[c]{@{}c@{}} Instructive \end{tabular} & 
\begin{tabular}{p{0.9\linewidth}|c}
\begin{tabular}{p{\linewidth}}
\texttt{What are the useful radiology descriptors to distinguish {\textrm{\emph{class\_names}}}?}
\end{tabular} & 62.06\\
\midrule
\begin{tabular}{p{\linewidth}}
\texttt{What are the helpful radiology descriptors to distinguish {\textrm{\emph{class\_names}}}?}
\end{tabular}
& \begin{tabular}[c]{@{}c@{}} 62.73 \end{tabular}\\
\midrule
\begin{tabular}{p{\linewidth}}
\texttt{What are the concise radiology descriptors to distinguish {\textrm{\emph{class\_names}}}?}
\end{tabular}
& \begin{tabular}[c]{@{}c@{}}63.27 \end{tabular}\\
\end{tabular}   \\
        \midrule
\begin{tabular}[c]{@{}c@{}} Misleading \end{tabular} & 
\begin{tabular}{p{0.9\linewidth}|c}
\begin{tabular}{p{\linewidth}}
\texttt{What are the irrelevant radiology descriptors to distinguish {\textrm{\emph{class\_name}}}?}
\end{tabular} & 52.96\\
\midrule
\begin{tabular}{p{\linewidth}}
\texttt{Give me some random visual features in a photo}
\end{tabular}
& \begin{tabular}[c]{@{}c@{}} 55.98 \end{tabular} \\
\end{tabular}
\\
\bottomrule
\end{tabular}}
\label{tab:robustness_check_prompts}
\end{table}

\section{Implementation Details}
\label{app:implementation_details}
\subsection{Datasets}
In this section, we provide details for creating datasets, pre-processing and train/val/test split for each dataset. 

First, for the curated four datasets with strong confounding factors: 

\textbf{(1) NIH-gender}: We use the meta data from NIH-CXR~\citep{wang2017chestx} to build a subset with male and female Chest X-rays for classifying Atelectasis and Effusion. The training set consists of male patients with Atelectasis and females with Effusion, and the test set consists of female patients with Atelectasis and males with Effusion. The confounding factor here is gender.

\textbf{(2) NIH-age}: We build a subset from NIH-CXR with young and old Chest X-rays to classify normal and abnormal. The training set consists of normal X-rays from young patients (age <= 16), and abnormal X-rays with from old individuals (age >= 70); the test set has the opposite combination. The confounding factor for this dataset is age. Normal and Abnormal are defined by the labels, normal X-rays are the ones with label ``No Findings", and abnormal are rest of the X-rays.

\textbf{(3) NIH-agemix}: Similar to NIH-age, for the normal training set, we have 90\% young patients and 10\% old paitents, for the abnormal training set, we have 90\% old patients and 10\% young patients. The test set has the same distribution as NIH-age, with normal X-rays from the old and abnormal X-rays from the young.

\textbf{(4) Covid-mix}: The Covid samples are a collection of open source releases, gathered on Kaggle, with 2,313 Covid X-rays in total~\footnote{\url{https://www.kaggle.com/datasets/amanullahasraf/covid19-pneumonia-normal-chest-xray-pa-dataset?select=covid}}. We build a balanced dataset by providing 2,313 pneumonia X-rays from two medical systems for training~\citep{kermany2018identifying} and testing~\cite{wang2017chestx}.
There can be several confounding factors in this dataset, for example, hospital, age and gender. 

Second, for the four standard benchmarks:

\textbf{(5) NIH-CXR}~\citep{wang2017chestx}: A large public dataset of chest X-rays, with 108,948 front view Chest x-ray images from NIH Clinical Center. Following previous setting~\citep{wang2022metateacher}, we use six main classes for classification, based on clinical importance and prevalence: Atelectasis, Cardiomegaly, Effusion, Consolidation, Edema and Pneumonia. 

\textbf{(6) Covid-QU}~\citep{chowdhury2020can}: A Covid dataset collected by Qatar University. The dataset consists of four classes, Covid-19, normal, lung opacity and pneumonia.

\textbf{(7) Pneumonia}~\citep{kermany2018identifying}: A public dataset for classifying pneumonia and normal, collected from Guangzhou Women and Children’s Medical Center. 

\textbf{(8) Open-i}~\citep{demner2012design}: Chest X-rays collected by Indiana University Hospital through the network from open source literature and biomedical image collection, with 3,955 radiology reports, corresponding to 7,470 frontal and lateral chest films. 
To be consistent with other datasets, we filter out the side chest x-ray in Open-i, leaving only frontal images. We further filter out diseases with few images, leaving three main classes: Cardiomegaly, Lung opacity, Normal.

Statistics of each dataset are in~\cref{tab:data_stats}. We report accuracy as the metric on all datasets.

\begin{table}[h!]
\centering
\caption{Statistics for the eight datasets.}
\begin{tabular}{lccccc}
\toprule
Dataset & Train/Val & Test & Dataset & Train/Val & Test \\ 
\midrule
NIH-gender  & 3000/500   & 1000   & NIH-CXR   &  8066/1140   & 2317  \\
NIH-age  & 4000/1000   & 1000   & Covid-QU     &  16930/2115 & 2114                     \\
NIH-agemix & 4000/1000 & 1000  & Pneumonia     &  5216/16  & 624                      \\ 
Covid-mix & 3700/462 & 462 & Open-i & 884/438 & 890 \\
          \bottomrule
\end{tabular}
\label{tab:data_stats}
\end{table}

\subsection{More implementation details}
For fair comparison, we use linear probing (a fully connected layer) to evaluate image features, Post-hoc and label-free CBM, and our method. We set the number of epochs to 5000 with early stopping. The batch size is set with a grid search in \{8, 16, 32, 64, 128\}, empirically we do not observe significant difference with difference batch sizes. The initial learning rate is set to 0.01 in all experiments with an Adam optimizer. 
We use average pooling of the heatmaps for all datasets, except for Covid-QU where adding max-pooling (local similarity) can improve performance. 
The datasets are all publicly available, though most of them need reading user agreements before using and directly sharing the data are not allowed. Experiments are done in a server with 8 NIVIDIA Titan RTX GPUs, training to get main results takes less than an hour for all datasets.
Our code is available anonymously at: 
\url{https://anonymous.4open.science/r/NeurIPS-submission-10334/}.

\section{Case Study and Visualizations}
\label{app:case_study}

\subsection{More Interpretation Examples} 
\begin{figure}[h!]
    \centering
    \includegraphics[width=0.49\linewidth]{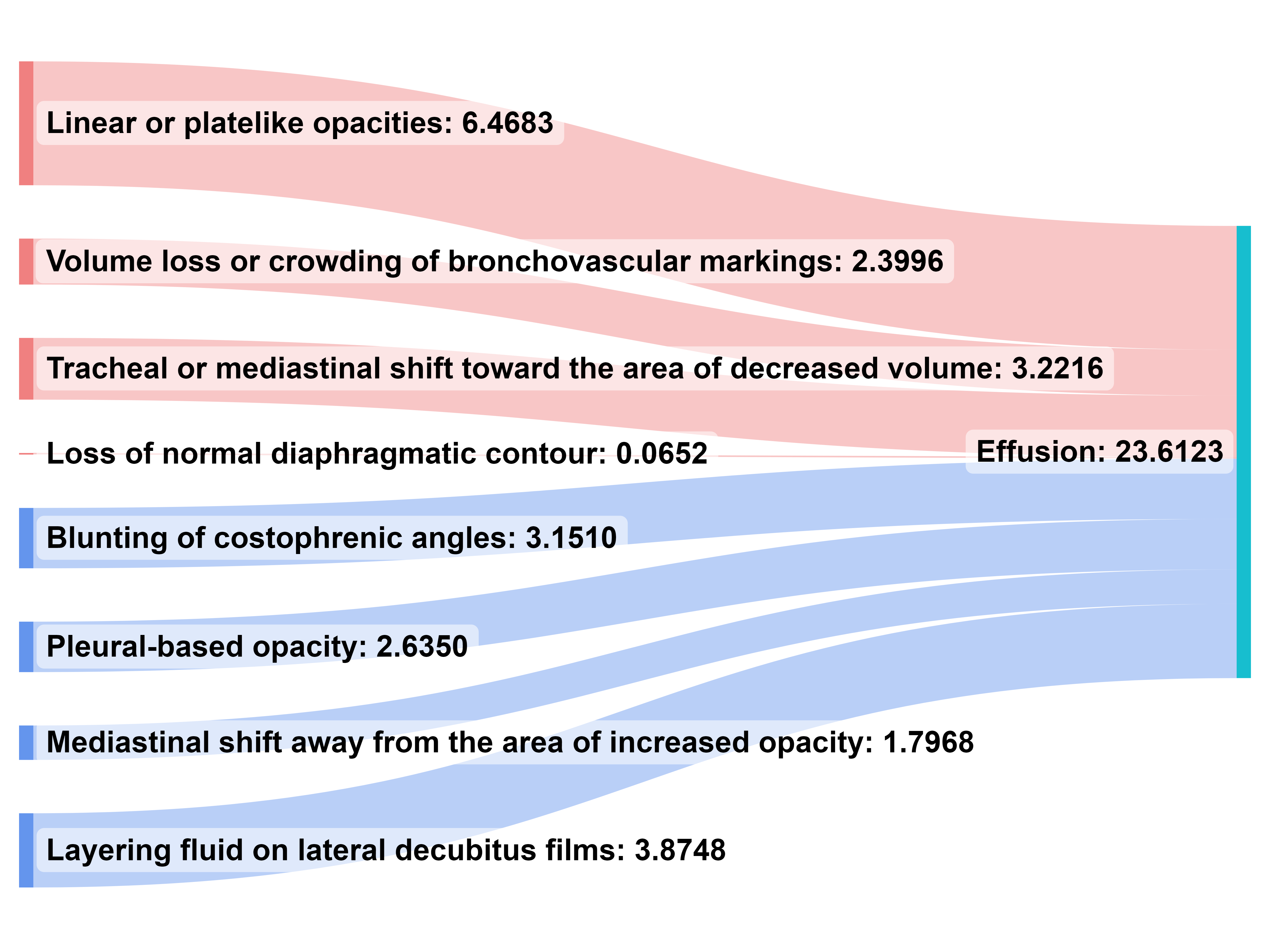}
    \hfill
    \includegraphics[width=0.49\linewidth]{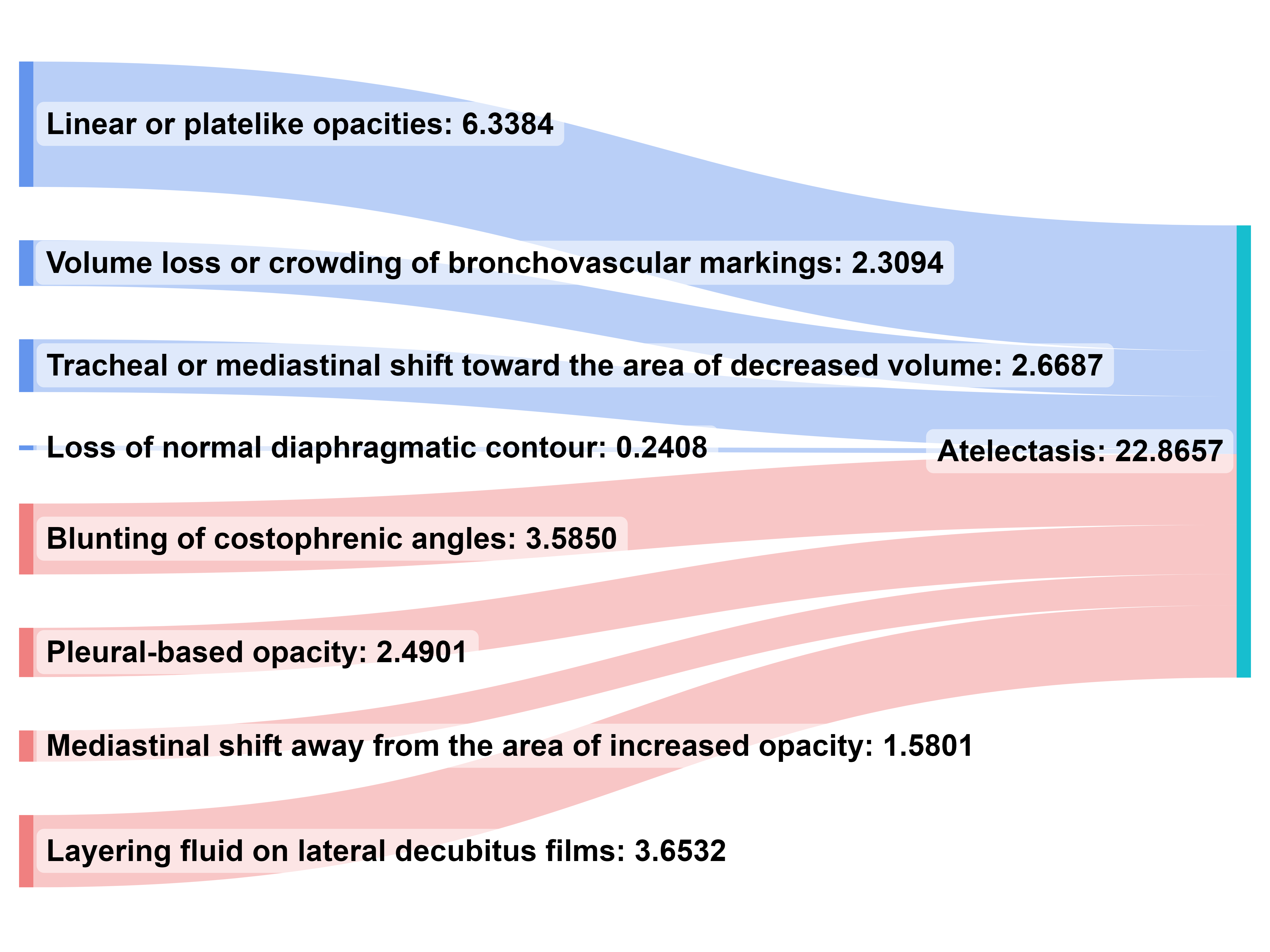}
    \caption{NIH-Gender GPT4 weights for Effusion and Atelectasis}
    \label{fig:nih_gender_gpt4_weights}
\end{figure}

We provide additional visualization of the learned final layer weights with a Sankey diagram of the final layer weights on the two classes for the nih-gender dataset, shown in~\cref{fig:nih_gender_gpt4_weights}.
The visualization also aligns with medical knowledge. For classifying Effusion, the model treat Effusion concepts as positive (e.g., ``Blunting of costophrenic angles'') and Atelectasis concepts as negative (e.g, ``Linear or platelike opacities''), and vice versa for classifying Atelectasis. 

\begin{figure}[h!]
\subfigure[NIH-gender Effusion Instances]
{\label{fig:nih_gender_effusion_instance}\includegraphics[width=0.475\linewidth]{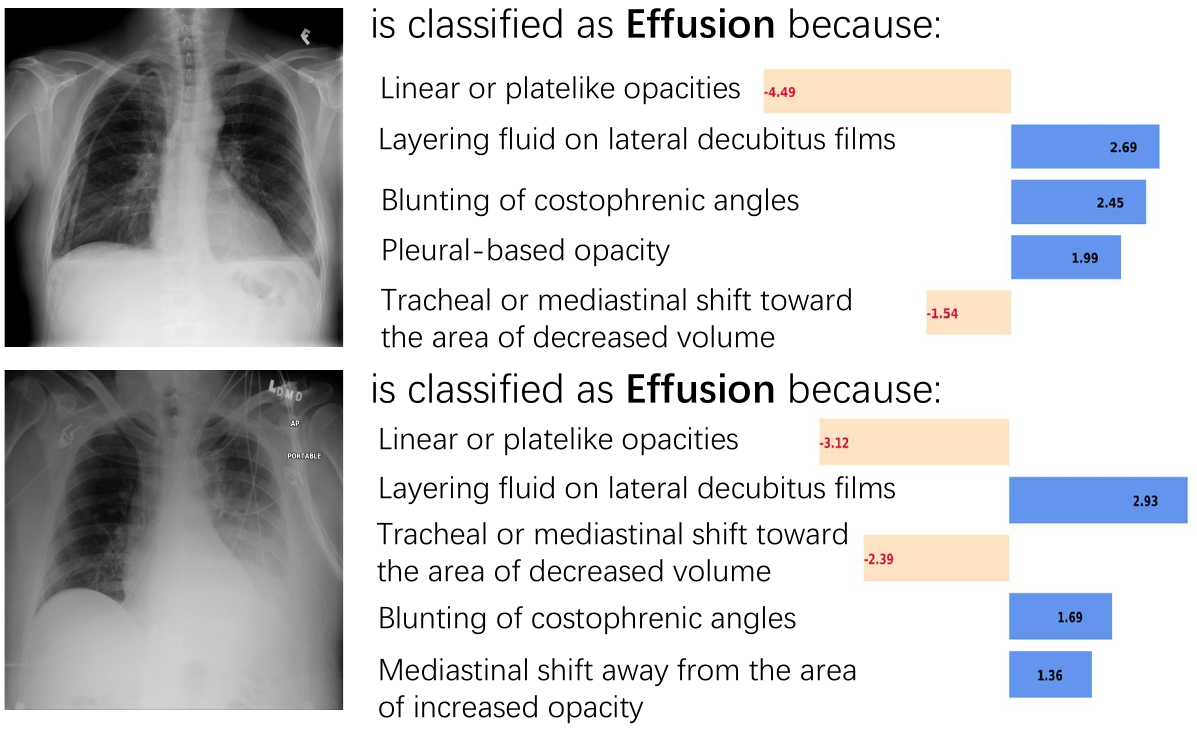}}
\subfigure[NIH-gender Atelectasis Instances]{\label{fig:nih_gender_atelectasis_instance}\includegraphics[width=0.475\linewidth]{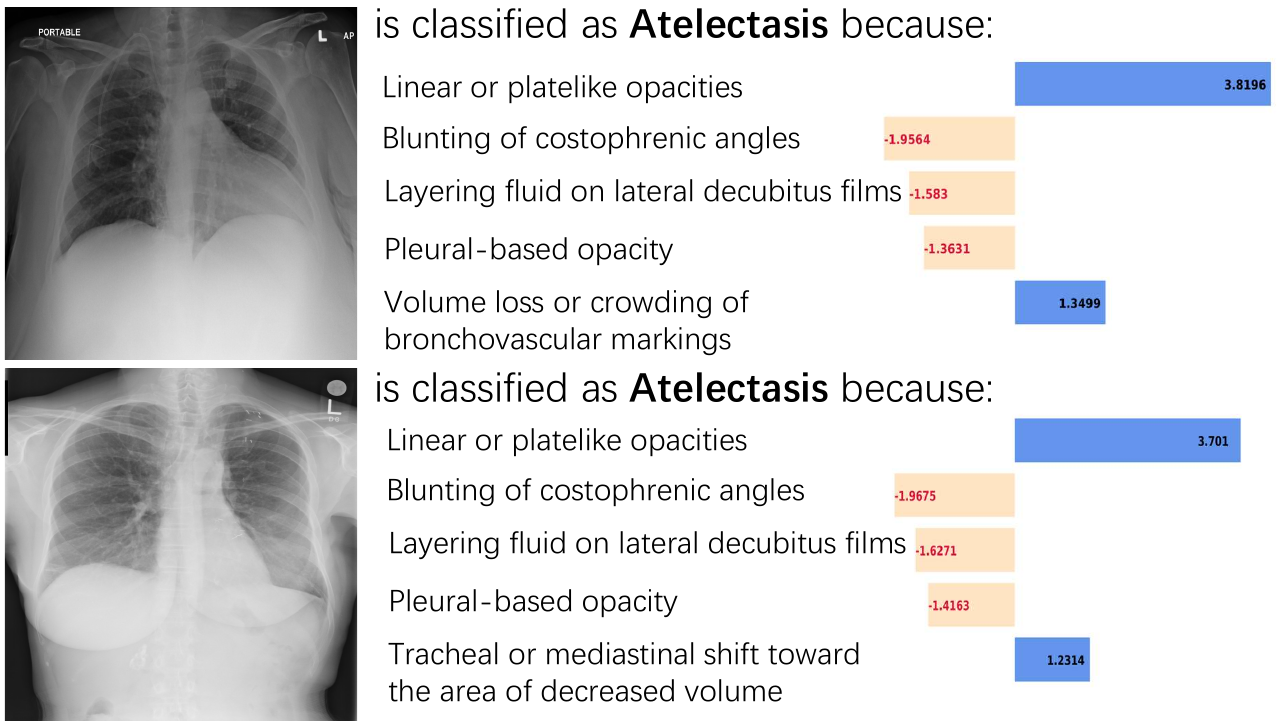}}
\caption{Examples of interpretability on dataset NIH-gender}
\vspace{-8pt}
\label{fig:nih_gender_instance_interpretability}
\end{figure}

Four examples of instance-level predictions from the NIH=gender dataset are visualized in~\cref{fig:nih_gender_instance_interpretability}. For X-rays with effusion, \textit{Linear or platelike opacities} is a important negative concept, while \textit{Layering fluid on lateral decubitus films} and \textit{Blunting of costophrenic angles} are the important positive concepts to make predictions. 

\subsection{Human Evaluation}
We further ask a board-certified radiologist to manually evaluate the quality of interpretations from our models. 
As manually reading X-rays are time-consuming even for experts, we ask the radiologist to carefully review ten Chest X-rays from the pneumonia datasets, where our model made correct predictions for these X-rays. The radiologist read each Chest X-ray, then evaluate if the instance-level interpretations (i.e., what are the important positive or negative concepts) make sense.
The radiologist mostly agree on the explanations produced by our model, e.g., consolidations and air bronchograms are the main observations from the Chest X-rays with pneumonia. Other than that, pleural effusion is hard to evaluate manually on recumbent films especially at low resolution.

\subsection{Error Analysis}
\begin{figure*}[h!]
\subfigure[X-rays misclassified Pneumonia Instances]
{\label{fig:chestxray_pneumonia_wrong}\includegraphics[width=0.475\linewidth]{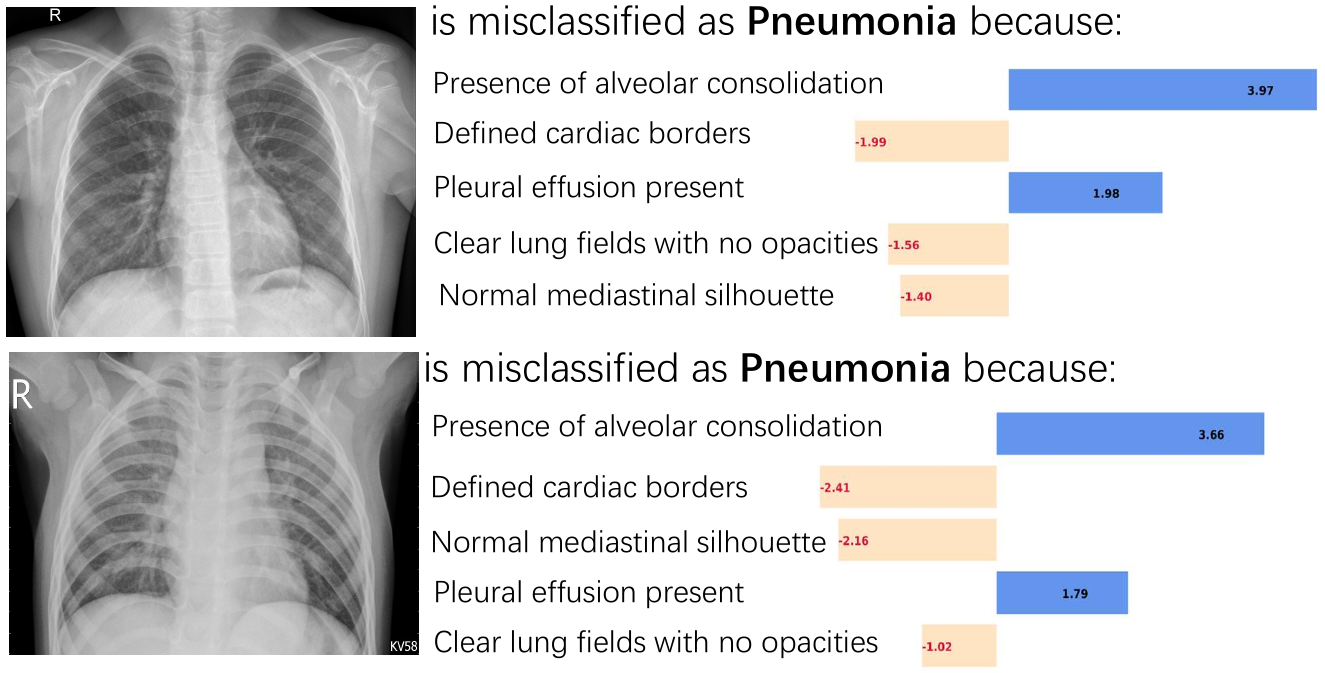}}
\subfigure[ChestXray Normal Instances]{\label{fig:chestxray_normal_wrong}\includegraphics[width=0.475\linewidth]{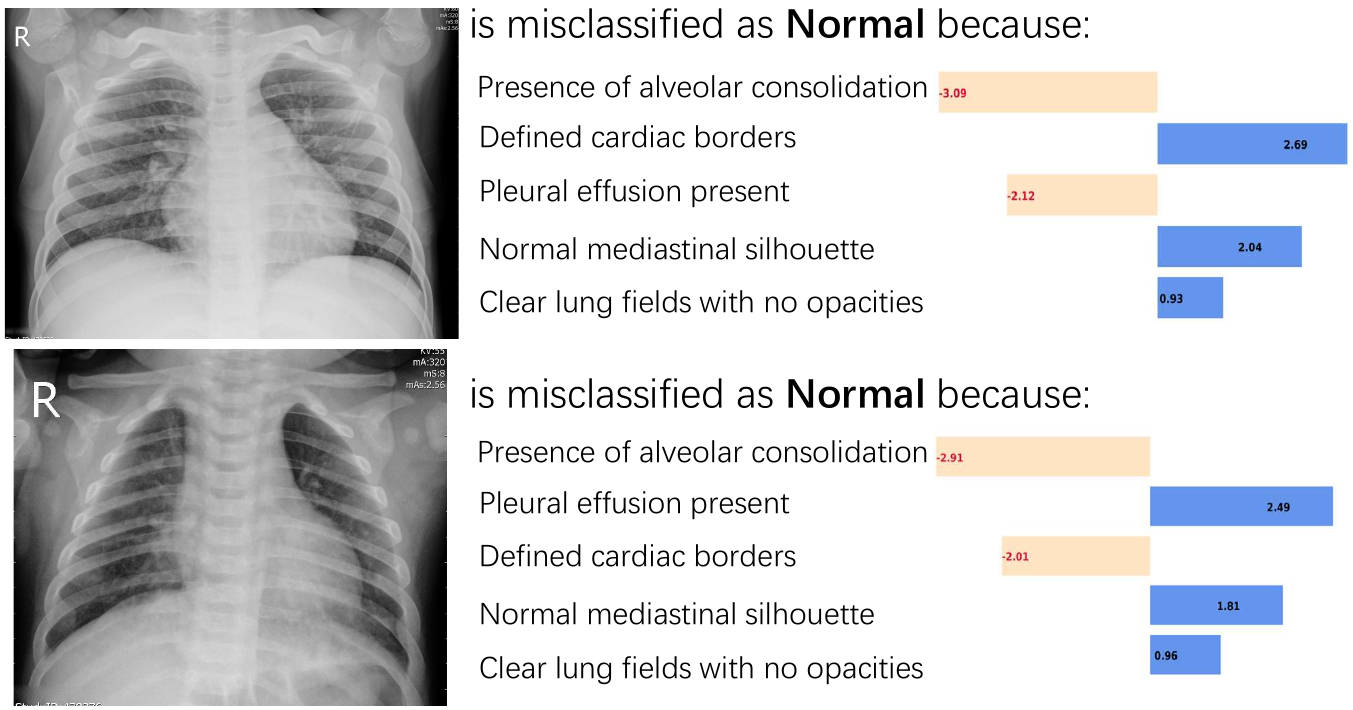}}
\caption{Error Analysis on the Pneumonia dataset}
\label{fig:chestxray_error}
\end{figure*}

We further analyze when the model would make wrong predictions, with four examples in~\cref{fig:chestxray_error}. On the pneumonia dataset, as the global interpretation of the model largely aligns with human, i.e., the linear weights could identify positive and negative concepts for a class, the main errors come from the wrong estimations of concept scores. The reason is our vision-language model BioViL is still not perfect to estimate the correlations between a concept and an image.
For example, a normal X-ray is mis-classified as pneumonia, because the model thinks this X-ray has high correlations with ``Presence of alveolar consolidation''.

\begin{figure}[h!]
    \centering
    \includegraphics[width=0.7\linewidth]{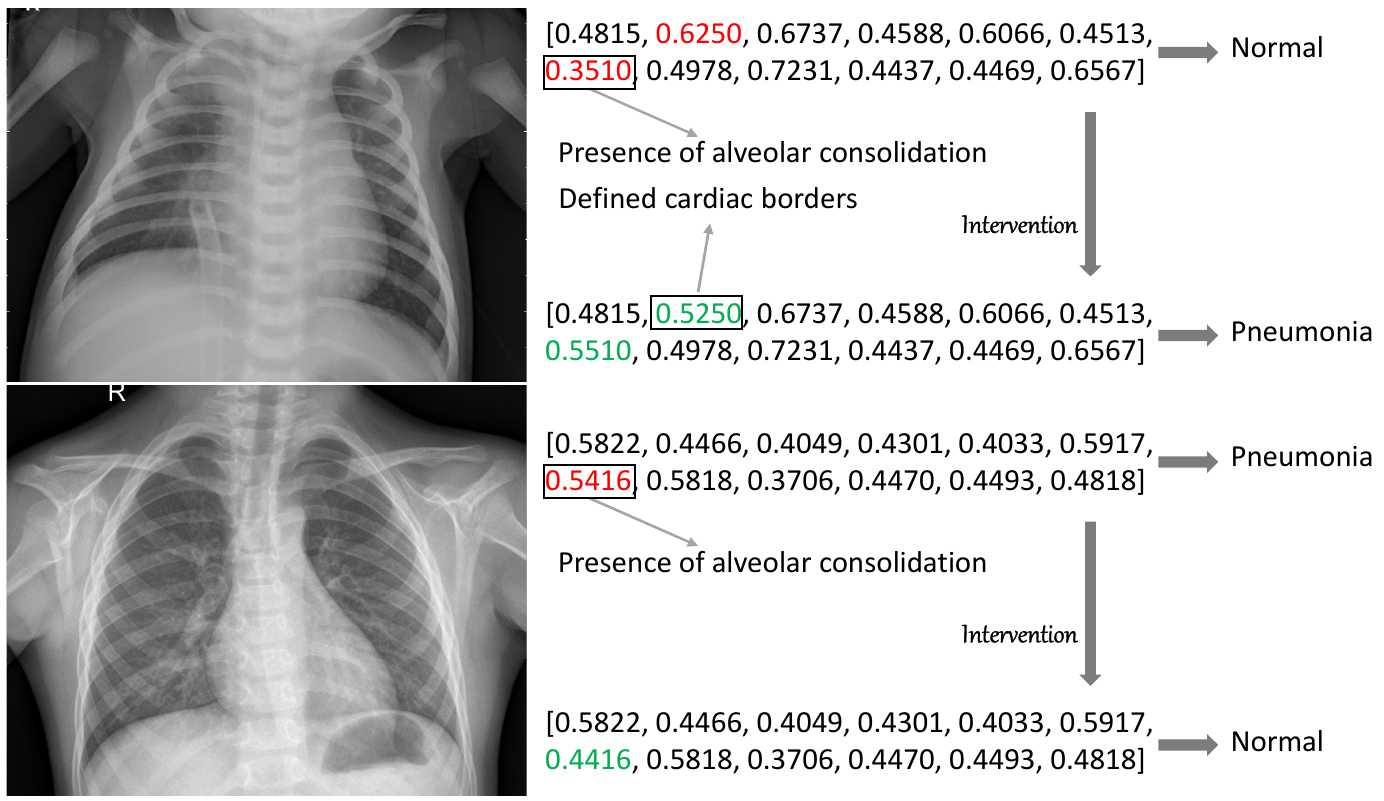}
    \caption{Intervention on the concept vector to correct the predictions.}
    \label{fig:chestxray_intervention}
\end{figure}

\paragraph{Test-time Intervention to correct the predictions} 
Based on the analysis, we present an advantage of our model, where we can do test-time intervention to correct model predictions. As analyzed above, once we noticed a wrong estimation by the VLM, we can correct the prediction by perturbing the wrong estimation. For example~(see~\cref{fig:chestxray_intervention}), if a normal X-ray is mis-classified as pneumonia because the VLM gives a high score to ``Presence of alveolar consolidation'', an expert can manually observe the error, then decrease the score to rectify the model's prediction.  Classification with concepts present a new opportunity for human-model interaction and diagnosing mistakes from neural models.  

\section{Additional Experiments}
\label{app:experiments}
\subsection{Comparison with other VLP models}
We compare the performance of BioViL with other vison-language models to demonstrate the importance of a good visual-language model for our task. 
Specifically, we conduct experiments for the comparison of BioViL over other VLP models: a non-medical domain VLM, CLIP~\cite{CLIP} and a medical domain VLM, MedCLIP~\cite{medclip}. 
We choose a strong model, CLIP ViT-L/14@336px as the CLIP encoder, and MedCLIP ViT as the MedCLIP model.
The results are shown in~\cref{tab:different_vlp_models_robust} and~\cref{tab:different_vlp_models_standard}. 
In general, a medical domain VLM would perform better than a non-medical domain VLM, especially when classifying with concepts, as the estimations of concept scores rely on a well-pretrained in-domain model.
\begin{table}[h!]
\vspace{-10pt}
    \caption{Performance comparison with other VLMs on robustness datasets.}
    \centering
    \begin{tabular}{cccccc}
    \toprule
         Models & NIH-gender & NIH-age & NIH-agemix & Covid-mix \\
        \midrule
        CLIP Visual Encoder & 21.30 & 4.00 & 14.40 & 49.57\\
        MedCLIP Visual Encoder & 21.80 & 5.50 & 16.70 & 52.81 \\
        BioViL Visual Encoder & 71.60 & 9.40 & 13.70 & 51.08 \\
        \midrule
        Ours with CLIP & 35.80 & 15.00 & 19.10 & \textbf{64.93}\\
        Ours with MedCLIP & 42.00 & 20.10 & 21.10 & 51.95 \\
        Ours with BioViL & \textbf{79.60} & \textbf{50.70} & \textbf{53.40} & 62.36\\
        \bottomrule
    \end{tabular}
    \label{tab:different_vlp_models_robust}
    \vspace{-5pt}
\end{table}

\begin{table}[h!]
\vspace{-10pt}
\caption{Performance comparison with other VLMs on standard datasets.}
    \centering
    \begin{tabular}{cccccc}
    \toprule
         Models & NIH-CXR & Covid-QU & Pneumonia & Open-i\\
        \midrule
        CLIP Visual Encoder &  48.16 & \textbf{93.03}  & 79.32 & 54.49 \\
        MedCLIP Visual Encoder & 48.55 & 89.02 & 81.73 & 52.58 \\
        BioViL Visual Encoder & \textbf{63.66} &  78.14 & 86.70 & 71.01 \\
        \midrule
        Ours with CLIP & 33.71 & 71.85 & 62.50 & 59.10 \\
        Ours with MedCLIP & 44.36 & 64.76 & 46.31 & 57.87 \\
        Ours with BioViL & 63.27 & 78.00 & \textbf{88.46} & \textbf{72.80} \\ 
        \bottomrule
    \end{tabular}
    \vspace{-10pt}
\label{tab:different_vlp_models_standard}
\end{table}

\subsection{Other operations on heatmaps}
As mentioned in~\cref{sec:method_clip}, though more advanced methods such as convolution on the heatmaps are possible, we find average pooling works well across all datasets, while preserving interpretability for the concept scores estimated by the VLM, shown in~\cref{tab:ablation_conv}. Future work could explore how to build local-aware VLMs and better leverage the spatial information to build robust models.

\begin{table}[h!]
\vspace{-10pt}
    \caption{Performance comparison for different operations on heatmaps.}
    \centering
    \begin{tabular}{cccccc}
    \toprule
         Models & NIH-gender & NIH-age & NIH-agemix & Covid-mix \\
        \midrule
        BioViL Image Features & 71.60 & 9.40 & 13.70 & 51.08 \\
        BioViL Image Features (dropouts) & 70.20 & 19.00 & 28.60 & 49.57 \\
        \midrule
        Convolution & 61.20 & \textbf{57.50}  & \textbf{56.60} & 50.00 \\
        Linear & 66.30 & 27.00  & 29.50 & 45.84 \\
        Avg pooling & \textbf{79.60} & 50.70 & 53.40 & \textbf{62.36} \\ 
        \bottomrule
    \end{tabular}
    \label{tab:ablation_conv}
    \vspace{-5pt}
\end{table}

\subsection{Training Curves}
\begin{figure}[h!]
    \centering
    \includegraphics[width=0.49\linewidth]{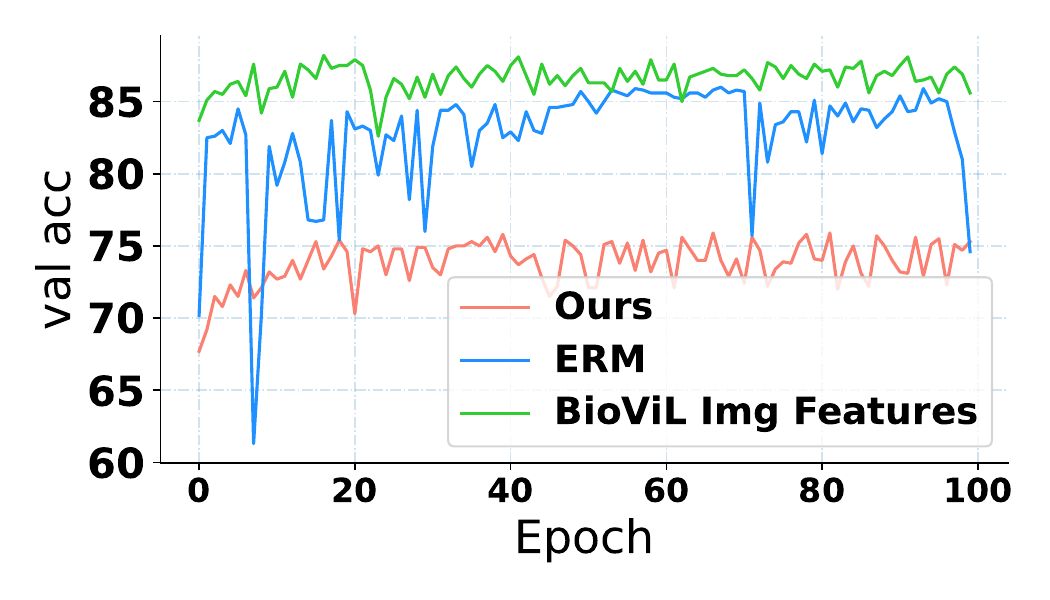}
    \includegraphics[width=0.49\linewidth]{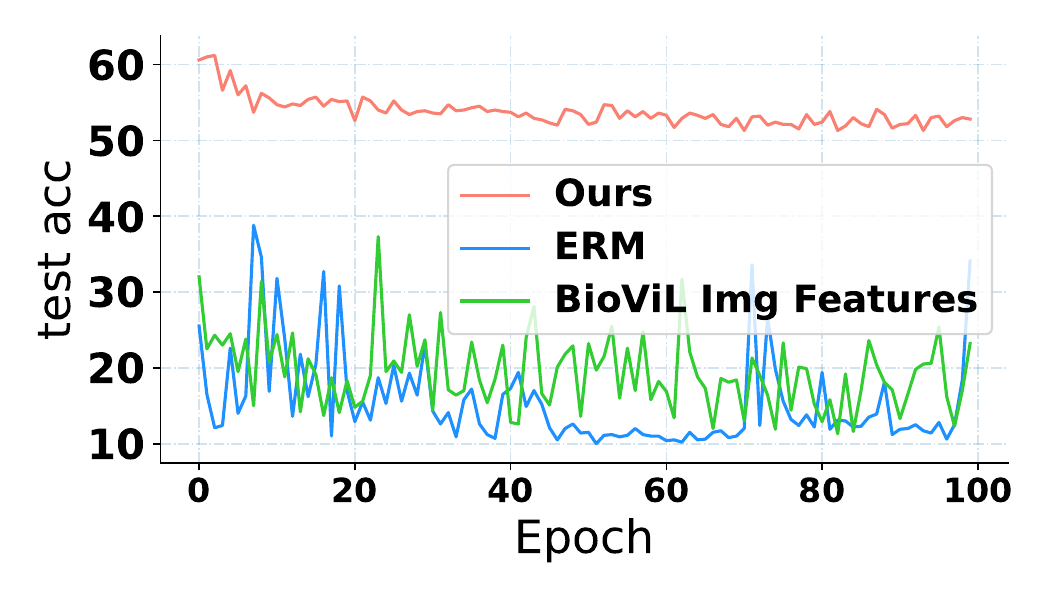}
    \caption{Validation and Test Accuracy along the training process on NIH-agemix}
    \label{fig:val_test_acc_wrt_epoch_nih_agemix}
\end{figure}

Similar to~\cref{fig:val_test_acc_wrt_epoch_nih_age}, we visualize the validation and testing accuracy on NIH-agemix in~\cref{fig:val_test_acc_wrt_epoch_nih_agemix}. ERM and BioViL Image Features struggle to learn the critical information for classification, but instead learning shortcuts of the data, i.e., predicting if the X-ray is from a young or old individual with high accuracy, but perform bad on the test set with different age groups from the validation set. 

\end{document}